\documentclass[preprint,12pt]{elsarticle}

\usepackage{amssymb}
\usepackage{amsmath}

\usepackage{comment}
\usepackage{threeparttable}
\usepackage[figuresright]{rotating}
\RequirePackage{pdflscape}
\usepackage{subfig}
\usepackage{amsmath}
\usepackage{algorithm}
\usepackage{algorithmicx}
\usepackage{algpseudocode}
\usepackage{apalike}
\usepackage{url}
\usepackage{hyperref}
\usepackage{xcolor}
\biboptions{authoryear}
\journal{Neural Networks}

\begin{document}

\begin{frontmatter}



\title{ST-Tree with Interpretability for Multivariate Time Series Classification}

\author[a,b]{Mingsen Du \footnote{This paper was completed by the first author when he was a MD student at
		Shandong Normal University and was supplemented during he was a Ph.D. student at Shandong 
		University.}} \ead{MingsenDu@163.com}
\author[b]{Yanxuan Wei} \ead{wyx\_num1@163.com}
\author[b]{Yingxia Tang} \ead{19854191173@163.com}
\author[b,c]{Xiangwei Zheng} \ead{xwzhengcn@163.com}
\author[a]{Shoushui Wei\corref{cor1}} \ead{sswei@sdu.edu.cn}
\author[b,c]{Cun Ji\corref{cor1}}  \ead{jicun@sdnu.edu.cn}

\address[a]{School of Control Science and Engineering, Shandong University, Jinan, China}
\address[b]{School of Information Science and Engineering, Shandong Normal University, Jinan, China}
\address[c]{Shandong Provincial Key Laboratory for Distributed Computer Software Novel Technology, Jinan, China}

\cortext[cor1]{Corresponding author}
\begin{abstract}
Multivariate time series classification is of great importance in practical applications and is a challenging task. However, deep neural network models such as Transformers exhibit high accuracy in multivariate time series classification but lack interpretability and fail to provide insights into the decision-making process. On the other hand, traditional approaches based on decision tree classifiers offer clear decision processes but relatively lower accuracy. Swin Transformer (ST) addresses these issues by leveraging self-attention mechanisms to capture both fine-grained local patterns and global patterns. It can also model multi-scale feature representation learning, thereby providing a more comprehensive representation of time series features. To tackle the aforementioned challenges, we propose ST-Tree with interpretability for multivariate time series classification. Specifically, the ST-Tree model combines ST as the backbone network with an additional neural tree model. This integration allows us to fully leverage the advantages of ST in learning time series context while providing interpretable decision processes through the neural tree. This enables researchers to gain clear insights into the model's decision-making process and extract meaningful interpretations. Through experimental evaluations on 10 UEA datasets, we demonstrate that the ST-Tree model improves accuracy in multivariate time series classification tasks and provides interpretability through visualizing the decision-making process across different datasets.
\end{abstract}


\end{frontmatter}



\section{Introduction}
\label{sec:intro}

Multivariate time series classification is an important and challenging task for time series, which are sequences of measurements collected from sensors. A time series is a set of data in chronological order and is widely used in a variety of domains and in a variety of real-world applications, including atrial fibrillation detection (\cite{li2023diagnosis}), error message detection (\cite{Yu2022LSTMBasedID}), intelligent device detection \cite{Du2023AutomaticCI}, sleep record classification \cite{Chambon2017ADL}, drunk driving detection (\cite{Li2015DrunkDD}), human activity recognition (\cite{Ma2019AttnSenseMA}) and so on.

Time series classification is one of the important tasks in time series analysis. Time series classification methods predict labels for unknown time series instances from the knowledge gained from existing labeled instances. In recent years, more and more researchers focus on the time series classification problem, which has important research value in the field of data mining (\cite{Du2024MultivariateTS, HUANG2023868, CHEN2021126}). High accuracy is a concern for all researchers. However interpretability is still important in many tasks. While much work has been done on interpretability in computer vision, tabular data classification and natural language processing, research on time series has not received as much attention (\cite{Rojat2021ExplainableAI}).

Most methods are not directly applicable to time series due to the different structure and nature of time series data (\cite{Ismail2020BenchmarkingDL}). In the early days of time series research, shapelets based methods were extensively studied (\cite{Ji2019AFS, Tao2022ProfilematrixbasedSD}). Shapelets provide interpretability of time series data and can be used to characterize time series by identifying and extracting representative shape patterns (\cite{Ruiz2020TheGM}). These shape patterns can be used to interpret the predictions of the model and thus enhance the understanding of the time series. However long time series data can cause the shapelet selection and extraction process to become more difficult, leading to increased computational complexity (\cite{Ji2019AFS}). There are also dictionary-based TSC methods (\cite{Middlehurst2019ScalableDC}), classification decisions can be explained by the distribution of symbolic words. However, this explanation is not intuitive enough because the distribution of symbolic words is difficult to judge directly. Some researchers can achieve high accuracy and interpretability through the combination of traditional shapelet and depth methods (\cite{Ji2022TimeSC, Ji2022FullyCN}). However, these methods are not intuitive for the classification decision process.

In recent years, a large number of deep model-based methods have been proposed, and commonly used models such as onvolutional Neural Network (CNN), Recurrent Neural Network (RNN), Fully Convolutional Network (FCN), Long Short-Term Memory (LSTM) and Transformer and so on (\cite{Wang2016TimeSC, Turbe2022InterpretTimeAN, FU2024106234, WAN2024106196}). Among them, Transformer is a deep neural network model based on the self-attention mechanism with strong time series modeling capability. Through the self-attention mechanism, Transformer can capture global dependencies and local patterns in time series data and extract rich feature representations. Deep models are widely used for their end-to-end and classification high accuracy. However, in numerous scenarios are in high-risk environments such as disease prediction or autonomous driving, the black-box characterization of these models is a significant drawback. Numerous approaches (\cite{Du2024MultivariateTS}) post-hoc interpretation such as gradcam, however, cannot reflect the decision-making process. Thus it becomes crucial to utilize some method to reflect the interpretability of the decision process of the deep model.

In multivariate time series classification, both deep neural network models and decision tree models have their advantages. However, two issues need to be addressed for these methods:  (1) \textbf{High accuracy}. Deep neural network models learn complex feature representations by (\cite{Ismail2020BenchmarkingDL}), resulting in high accuracy rates. Various methods based on deep models have emerged. The vast majority of deep models are black-box, post hoc explanatory models. (2) \textbf{Interpretability}. The decision tree model is interpretable and can clearly show the decision-making process, making the results easier to understand and interpret (\cite{Theissler2022ExplainableAF,Kim2022ViTNeTIV}). Both deep neural network models and decision tree models have their own advantages and limitations. In order to balance accuracy and interpretability, researchers have proposed methods that combine both (\cite{Pagliarini2022NeuralSymbolicTD}). 

\begin{figure}[h]
	\centering
	\includegraphics[width=0.7\textwidth]{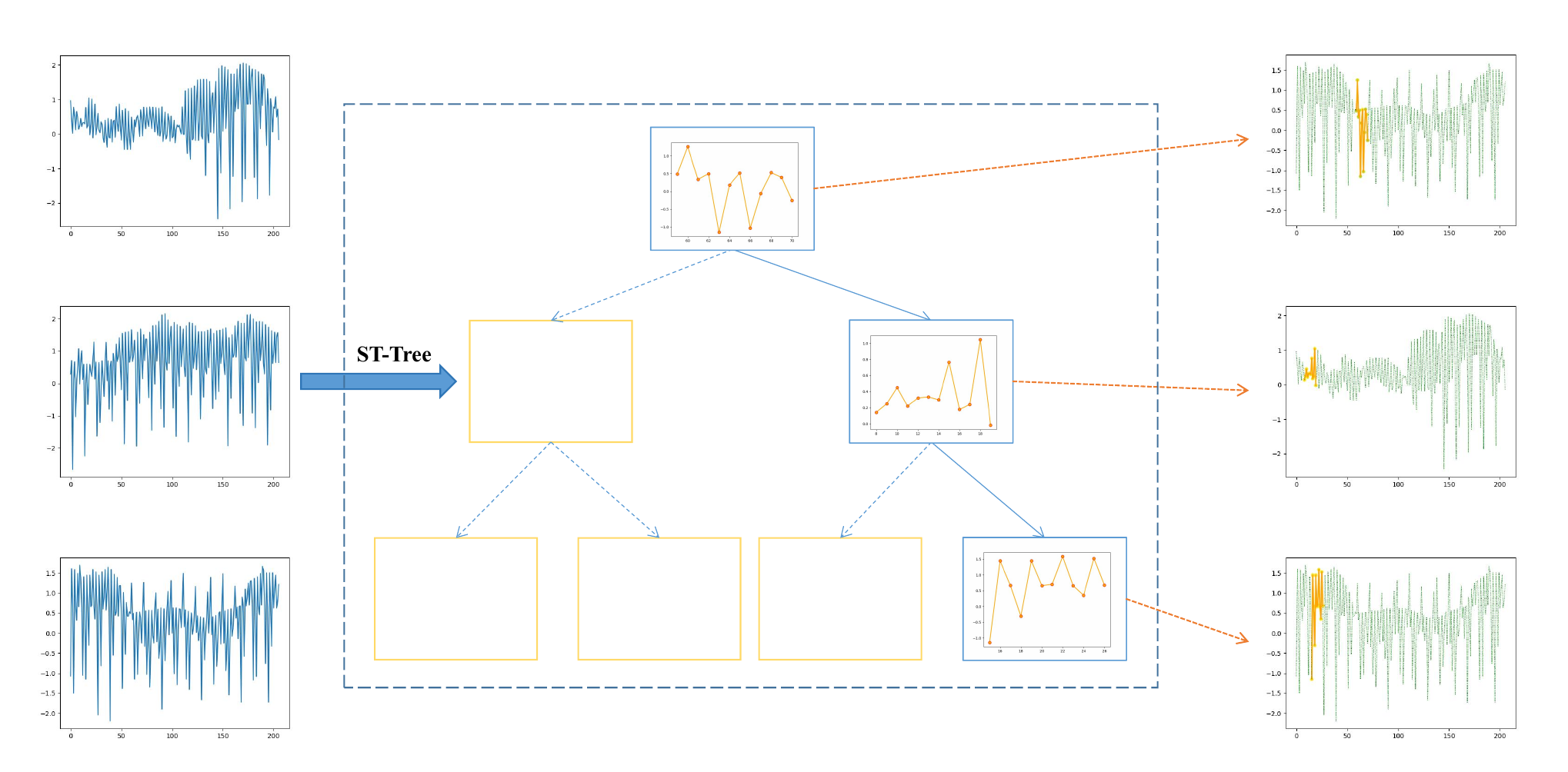}
	\caption{ST-Tree's transparent decision-making process, with the original Epilepsy time series on the left, the tree structure in the middle, and the positions of the prototypes corresponding to the original time series is on the right.}
	\label{example}
\end{figure}

In response to the above challenges, we proposed ST-Tree with interpretability for multivariate time Series classification to deal with the problem of model interpretability while balancing high accuracy and an interpretable decision process. The method consists of two parts: the time patch module and the neural tree module. Specifically, the time patch module uses ST as a feature extractor and time patch generator. To improve the interpretability of the model and visualize the decision-making process, the generated time patches are fed into a tree model, the neural tree module, which hierarchically paths the multivariate time series through a binary tree. Finally, the leaf prediction module perform label prediction on the multivariate time series by calculating branch routing scores.
The transparent decision process of ST-Tree is shown in Fig.~\ref{example}, each node corresponds to a time series segment closest to the prototype.
The main contributions of this paper are as follows:
\begin{enumerate}
	\item We propose the ST-Tree model, which combined ST with an additional neural tree model, which can take full advantage of ST to learn the time series context, and can use neural tree to provide an interpretable decision process.
	
	\item The proposed neural tree model decision-making itself is transparent, and each node judgment criterion can be clearly understood through the prototypical solution of the nodes.
	
	\item The ST-Tree model is demonstrated to have improved accuracy on multivariate time series classification tasks through experimental evaluation on 10 UEA datasets.
	
	\item We visualize the tree model on different types of datasets and illustrate the good interpretability of the model through the decision-making process.
\end{enumerate}

The rest of this paper is organized as follows: Section 2 provides an overview of the related work in the field. In Section 3, we present our method in detail. Section 4 describes the sensitivity experiments conducted to validate our approach, comparing it with other state-of-the-art methods. Finally, Section 5 concludes the paper.

\section{Related work}

\subsection{Transformer based methods}
\cite{Zuo2023SVPTAS} introduced a variable position coding layer and a self-attention mechanism based on variable-position to utilize variable and positional information of shapes.
\cite{Cheng2023FormerTimeHM} proposed FormerTime, which uses a hierarchical network with multiscale feature maps, an efficient temporally-approximate attention layer, and a contextual positional encoding strategy.
\cite{Du2023MultifeatureBN} proposed MF-Net, which acquires local features through an attention block, global features through a sparse self-attention block, and integrates them using a graph neural network (GNN) to capture spatial features.
\cite{Yang2023DyformerAD} presented Dyformer, which decomposes time series into subsequences with different frequency components, employs adaptive learning strategies based on dynamic architecture, and introduces a feature map attention mechanism to capture multi-scale local dependencies.
\cite{Hllig2022TSEvoEC} introduced TSEvo, which combines different time series transformations to generate realistic counterfactuals using an integrated time series Transformer.
 \cite{Du2024MultivariateTS} proposed MTSC\_FF, which utilizes a continuous wavelet transform-based attention layer to extract frequency domain features. It also employs a sparse self-attention layer to extract long-range features and captures spatial correlation using Kendall coefficients.

\subsection{Methods with interpretability}

\subsubsection{Traditional methods with interpretability }
\
\cite{Chen2023LocalizedSS} propose an interpretable approach for selecting time shapelets that incorporates position and distance metrics to assess the discriminative ability of candidate shapes, enhancing interpretability through the shape transformation process.
\cite{Manzella2021IntervalTR} proposed algorithms for extracting Interval Temporal Logic Decision Trees and extend the approach to Interval Temporal Random Forests, maintaining logical interpretability while modeling patterns at the propositional level.
\cite{Nguyen2019InterpretableTS} introduced an algorithm that combines symbolic representations at multiple resolutions and domains, leveraging extended symbol sequence classifiers to filter the best features using a greedy selection strategy. 
\cite{Hou2016EfficientLO} employed a generalized feature vector approach and a fused lasso regularizer to learn shape locations and obtain sparse solutions, outperforming previous shape-based techniques in terms of speed. \cite{Delaney2020InstanceBasedCE} proposed Native-Guide for generating proximal and reasonable counterfactuals for instance-based time series classification tasks by leveraging local in-sample counterfactuals. 
\cite{Bahri2022ShapeletbasedTA} introduced rule transform (RT) for generating discriminative temporal rules. RT creates a new feature space representing the support for temporal rules using Allen interval algebra and shapelets as units. 
\subsubsection{Deep methods with interpretability }
\
\cite{Younis2023FLAMES2GraphAI} introduced FLAMES2Graph, which visualizes highly activated subsequences and captures temporal dependencies using an evolutionary graph to explain deep learning decisions.  \cite{Lee2023ZTimeEA} proposed Z-Time, utilizing temporal abstraction and relationships to create interpretable features across multiple time series dimensions. \cite{Tang2019FewshotTC} proposed DPSN, which trains a neural network-based model and interprets it using representative time-series samples and shapes. Dual-prototype shapes are generated, representing the overall shape and discriminative partial-length shapes of different classes. \cite{Pagliarini2022NeuralSymbolicTD} presented a hybrid approach combining neural networks and temporal decision trees to capture temporal patterns and make decisions based on temporal features. Ma et al. \cite{Ma2020AdversarialDS} introduced ADSN, which employs an adversarial training strategy to generate shapelets that closely resemble the actual subsequence. Shapelet transformation is then used to extract discriminative features. \cite{Wan2023MemorySL} proposed an early classification memory shapelet learning framework that utilizes an in-memory distance matrix and optimizes for accuracy and earlyness to extract interpretable shapes. \cite{Zhu2021NetworkedTS} presented a networked time series shapelet learning approach that incorporates spatio-temporal correlations using a network impedance-based adjacency matrix. \cite{Ghods2022PIPPI} proposed PIP, an interpretable method that co-learns classification models and visual class prototypes, providing a higher combination of accuracy and interpretability. \cite{Fang2018EfficientLI} discovered shape candidates, measure their quality using coverage, and adjust them using a logistic regression classifier.

\section{Method}


\begin{table}[h]
	\centering
	\caption{Parameters}
	\label{tabparameters}
	\begin{tabular}{cl}
		\hline
		\textbf{Parameter} & \textbf{Meaning} \\ 
		\hline
		$d$ & Tree depth, representing the maximum depth of the tree structure. \\ 
		$\theta$ & Train parameters associated with the ST module. \\ 
		$\omega$ & Train parameters associated with the tree module. \\ 
		$\mathcal{P}$ & Prototype set, a collection of prototypes for similarity calculations. \\ 
		$T$ & Initialized tree. \\ 
		$L$ & Leaf node set, containing the IDs of the leaf nodes in the tree. \\ 
		$C $& Number of MTS channels dimension.\\
		$L'$ & Length dimension of MTS. \\
		$c^a$ & Convolution operation with filter size $a$. \\
	    $i$ & Current node index being traversed in the tree. \\ 
		$\mathbf{z}_i$ & Input feature vector associated with node $i$. \\ 
		$\mathcal{N}(\mathbf{z}_i)$ & Maximum similarity score between feature vector $\mathbf{z}_i$ and prototypes in $\mathcal{P}$. \\ 
		$\mathcal{R}_{i,j}(\mathbf{z}_i)$ & Routing score from node $i$ to child node $j$, indicating the branching direction. \\
		 
		
		$\mathcal{T}$ & Training set used for training the model. \\ 
		$E$ & Number of epochs for training. \\ 
		$f$ & ST module function used for feature extraction. \\ 
		$B$ & Number of batches. \\ 
		
		
		$\mathbf{z}_0$ & Output time patch of the ST module. \\ 

	
		$\hat{y}$ & Final predicted class label from the ST-Tree module. \\ 
		$N(\cdot)$ & Set of nodes in the binary tree. \\ 
		$E_{i,j}(\cdot)$ & Edge connecting parent node $i$ to child node $j$. \\ 
		$\mathbf{\tilde{z}}$ & Reshaped time patch used for similarity calculations. \\ 
		$\mathcal{E}_{i,j}(\mathbf{z}_i)$ & Function that transforms input feature $\mathbf{z}_i$ for node $i$ and child node $j$. \\ 
		$\mathbf{\rho}$ & Vector of routing scores for each node, representing similarity with prototypes. \\ 
		$x$ & In this study, x is used as an intermediate variable for time patch calculations. \\ 
		\hline
	\end{tabular}
\end{table}

\begin{figure*}
	
	\centering
	\includegraphics[width=0.95\textwidth]{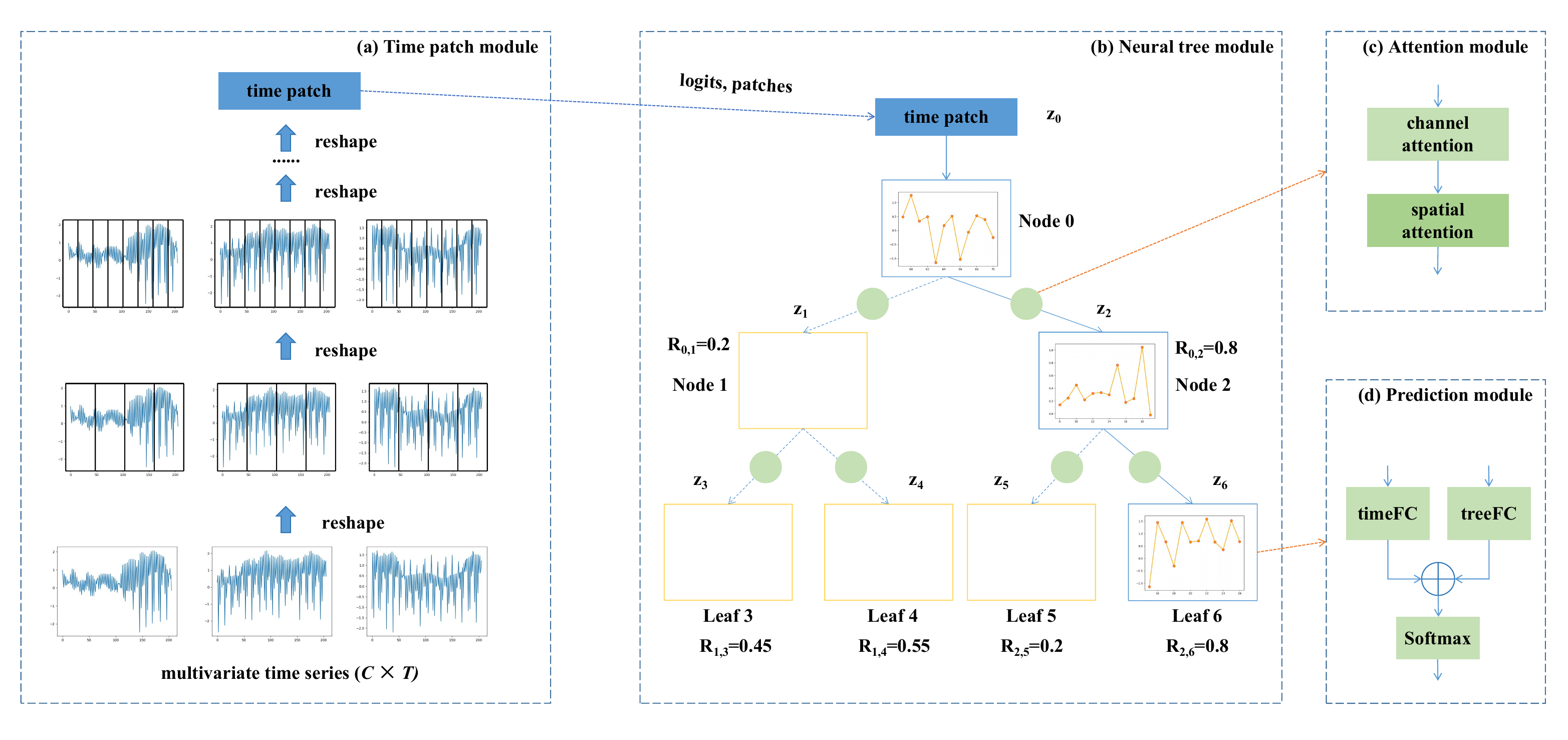}
	\caption{Schematic of ST-Tree classification via tree structure, with the original Epilepsy time series on the left, the tree structure in the middle. The nodes in the tree compute routing scores on the basis of the prototypes. In the figure, \( z_{0} - z_{6} \) represent the prototype of each node. \( R_{0, 1} - R_{2, 6} \) represent the probability of the left or right child node.}
	\label{overview}
\end{figure*}
ST (\cite{Liu2021SwinTH}) brings certain advantages for processing multivariate time series. It employs a patch-based processing strategy, dividing the time series data into parallelizable sub-sequences known as time patches, similar to patches in images. Additionally, by incorporating self-attention mechanism, it becomes possible to model the contextual information between different time steps in the series, enabling the capturing of temporal relationships and dependencies.

To leverage these advantages, we propose the ST-Tree, illustrated in Fig.~\ref{overview}. It utilizes a time patch module (shown in Fig.~\ref{overview} (a)) to extract time patches. These time patches are then fed into the prototype-based neural tree module (depicted in Fig.~\ref{overview} (b)), which enhances the interpretability of the model.

\subsection{The pseudocode to train ST-Tree}
Algorithms \ref{algtree} and \ref{algtraverse} describe the training process for ST-Tree.

\begin{algorithm} 	
	\caption{Training Process for ST-Tree}
	\label{algtree}
	\footnotesize
	\begin{algorithmic}[1]
		\Require Training set $\mathcal{T}$, number of epochs $E$, ST module $f$, ST parameters $\theta$, tree model parameters $\omega$
		
		\For{$e \in\{1, \ldots, E\}$}
		\State Split $\mathcal{T}$ into $B$ batches $\{\mathcal{T}_1, \ldots, \mathcal{T}_B\}$ \Comment{Batch splitting process}
		
		\For{batch $b \in \{1, \ldots, B\}$}
		\State Extract batch $\left(x_b, y_b\right)$ from $\mathcal{T}_b$ 
		
		\State $\mathbf{z}_0^b \gets f(x_b; \theta)$ \Comment{Output from the ST module}
		
		\State  $\hat{y}_b \gets \Call{traverse}{(1, \mathbf{z}_0^b; \omega)}$ \Comment{Traverse the tree to get predictions}
		
		\State  $\text{loss}(\hat{y}_b, y_b)$ \Comment{Calculate the loss based on predictions}
		
		\State Update $\theta$ and $\omega$  \Comment{Optimize model parametersusing Adam optimizer}
		\EndFor
		\EndFor 
		
	\end{algorithmic}
\end{algorithm}

\subsection{Time patch module} \label{timepatchsec}
\begin{figure*}
	\centering
	\includegraphics[width=0.9\textwidth]{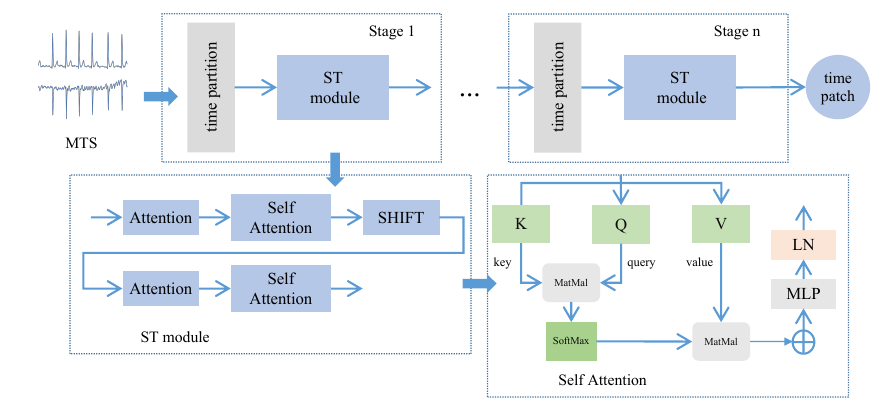}
	\caption{Schematic of ST module.}
	\label{timepatch}
\end{figure*}

We optimize a ST-based architecture, specifically the time patch module, for efficient feature extraction and obtaining time patches. The structure of the time patch module is illustrated in Fig.~ \ref{timepatch}. Time partition module is utilized as the first step to segment and obtain the initial time patch. The multivariate time series $X$ of dimensions $(T \times C)$ is reshaped using four non-overlapping neighboring timestamps to generate a time patch, which is embedded with dimensions $\frac{T}{4n} \times 4nC$. The attention module (detailed in Section ~\ref{att}) is employed as the second step for feature refinement. Embedding the attention module into the self-attention structure helps it to obtain more differentiated time domain features and time patches. Finally, the features pass through a Layer Normalization (LN) layer and an Multi-Layer Perceptron (MLP) layer for further processing.

The self-attention calculation formula is shown in Eq.~\eqref{eq1}. The query ($Q$), key ($K$) and value ($V$) are the matrices by extracting features from attention module (detailed in Section \ref{att}). 

\begin{equation} \label{eq1}
	X_{SA}=softmax\left(\frac{attention _{Q} X\left( attention _{K} X\right)^{T}}{\sqrt{d_{k}}}\right)  attention_{V} X
\end{equation}

\textbf{SHIFT module.} It incorporates a shift-window mechanism to limit the self-attention computation to non-overlapping local temporal patch windows. This design choice enables the interaction of attention between individual time patches.

\subsection{Neural Tree module}
\subsubsection{Overview} 

	After obtaining patches from the Time Patch Module in Section \ref{timepatchsec}, the next steps involve initializing a tree.
	\begin{enumerate}

	\item{Feature Transformation}
	
	First, the input \(\text{patches}\) are transformed using a sigmoid activation as Eq. \ref{feature_transformation}, where
	\(\text{patches}\) is the input feature tensor of ST model representation in Section \ref{timepatchsec}.
	
	\begin{equation}
		\text{patches}' = \text{sigmoid}(\text{patches})
		\label{feature_transformation}
	\end{equation}


	\item{Prototype Chunking}
	
	Then, the prototype vector \(\text{prototypes}\) is chunked into multiple prototypes as Eq. \ref{prototype_chunking}, where \(\text{prototypes}\) is the prototype tensor,~
	\(\text{num\_prototypes}\) is the number of prototypes to chunk. These prototypes will correspond to each node in the tree.
	
	\begin{equation}
		\text{chunked\_prototypes} = \text{chunk}(\text{prototypes},~\text{num\_prototypes})
		\label{prototype_chunking}
	\end{equation}

	\item{Tree Traversal}
	
	Finally, the transformed features and prototypes are processed through the tree structure. For each node in the tree, the output is computed as Eq. \ref{tree_traversal}, where
	\(\text{logits}\) is the prediction of ST model.
	\(\text{patches}'\) is the transformed feature tensor obtained from \(\text{feature\_transformation}\).
	 The out is the final prediction result.
	
	\begin{equation}
		\text{out} = \text{TreeTraversal}(\text{logits},~ \text{patches}',~\text{chunked\_prototypes})
		\label{tree_traversal}
	\end{equation}

	\end{enumerate}

\subsubsection{Recursive Tree}


Let \( T(i, d) \) be the tree initialization function where \( i \) is the index and \( d \) is the current depth. The formula for tree traversal is given in Eq. \ref{eqtree_initialization}, with detailed steps outlined in Algorithm \ref{algtraverse}.

\begin{equation}
	T(i, d) =
	\begin{cases}
		\text{Leaf}(i, \text{num\_classes}) & \text{if } d = \text{tree\_depth} \\
		\text{Branch}(i, \, T(i + 1, d + 1), \, \\ 
		T(i + \text{left.size} + 1, d + 1), \, \text{prototypes}) & \text{if } d < \text{tree\_depth}
	\end{cases}
	\label{eqtree_initialization}
\end{equation}

where
\text{Leaf}(i, \text{num\_classes}) generates a leaf node with index \( i \) and number of classes when \( d \) reaches the maximum \text{tree\_depth}.
\text{Branch} creates a branch node when \( d < \text{tree\_depth} \).
\( \text{left.size} \) is the size of the left subtree, used to determine the starting index of the right subtree.
\( \text{prototypes} \) represents the current prototypes for the branch node.

\subsubsection*{Pseudocode for Traverse Tree through Recursive}

\begin{algorithm} [H]	
	\caption{Traverse a Tree with Routing Scores}
	\label{algtraverse}
	\footnotesize
	\begin{algorithmic}[1]
		
		\Require Tree depth $d$, ST model parameter $\theta$, tree parameter $\omega$, prototype set $\mathcal{P}$ 
		\Ensure A traversed tree with routing scores
		
		\State $T \gets \text{init\_tree}(d, \omega)$ 
		\Comment Initialize tree with depth $d$ and parameter $\omega$
		
		\State Leaf node set $L \gets \emptyset$ 
		\Comment Initialize the leaf node set as empty
		
		\State $L \gets \text{assign\_leaf\_nodes}(T)$ 
		\Comment Assign node IDs of tree $T$ to the leaf node set $L$
		
		\Function {traverse}{$i, \mathbf{z}_{i}$}
		\Comment Recursively traverse the tree, starting from node $i$ and input feature $\mathbf{z}_{i}$
		
		\If{$i \in L$}
		\Comment If node $i$ is a leaf node
		\State \Return $\mathcal{L}(\mathbf{z}_{i})$ 
		\Comment Return class probabilities obtained through FC layers
		\EndIf
		
		\State $\mathcal{N}(\mathbf{z}_{i}) \gets \max(\text{similarity}(\mathbf{z}_{i}, \mathcal{P}))$
		\Comment Calculate the maximum similarity between node $i$'s feature $\mathbf{z}_{i}$ and the prototype set $\mathcal{P}$
		
		\State $\mathcal{R}_{i, 2i}(\mathbf{z}_{i}) \gets \mathcal{N}(\mathbf{z}_{i})$
		\Comment Routing score for the left child node (index $2i$)
		
		\State $\mathcal{R}_{i, 2i+1}(\mathbf{z}_{i}) \gets 1 - \mathcal{N}(\mathbf{z}_{i})$
		\Comment Routing score for the right child node (index $2i+1$)
		
		\State $l_{\text{child}} \gets \Call{traverse}{2i, \mathcal{E}_{i, 2i}(\mathbf{z}_{i})}$
		\Comment Recursively traverse the left child node
		
		\State $r_{\text{child}} \gets \Call{traverse}{2i+1, \mathcal{E}_{i, 2i+1}(\mathbf{z}_{i})}$
		\Comment Recursively traverse the right child node
		
		\State \Return $\mathcal{R}_{i, 2i}(\mathbf{z}_{i}) \times l_{\text{child}} + \mathcal{R}_{i, 2i+1}(\mathbf{z}_{i}) \times r_{\text{child}}$
		\Comment Return the weighted sum of the left and right child's routing scores
		\EndFunction
		
	\end{algorithmic}
\end{algorithm}

\subsection{Neural tree module}

\
\newline ST-Tree uses prototypes to identify discriminative regions in time patches. The prototype feature vector serves as a representative of a time series category. When discriminative features are sampled, they are routed to child nodes based on a routing direction determined by a routing module. Specifically, each tree node corresponds to a trainable prototype $\mathcal{P}_i$ 
for measuring the routing score $\mathcal{N}_i \in [0,1]$ for the reshaped time patch $\mathbf{z}$, 
as shown in Fig.~\ref{overview}(b).

We introduce a module that uses perfect binary trees and a routing mechanism. The module consists of sets of nodes $N(\cdot)$, leaf nodes $L(\cdot)$, and edges $E_{i,j}(\cdot)$ connecting parent node $i$ to child node $j$. This module represents a perfect binary tree, where each internal node has two children: $N_{2 \times i}$ and $N_{2 \times i+1}$. Given the encoder output $\mathbf{z_0} = f(x;\theta)$, the module predicts the final class label $\hat{y}$ using a soft decision-making process.

\subsubsection{Branch routing module}


\begin{enumerate}
	\item	
\textbf{Computing L2 Distance for Similarity} \label{seccomputefp}

ST-Tree computes the squared Euclidean distance between the time patch and the prototype of each category (as described in Eq. \ref{square_operation}-\ref{eqsimilarity}). This distance is considered as the similarity score between the time patch and the category prototype, where a smaller distance value indicates a higher similarity score. To quantify the similarity between the prototype and the time patch in each node, the similarity score is determined as in (\cite{Chen2018ThisLL}).

The distance is measured using a convolutional operation, where each prototype acts as a convolutional kernel on $\mathbf{z}$. The distance between the prototype and $\mathbf{z}$ is computed, and we represent the similarity by using a logarithmic similarity measure as briefly defined in Eq.~\eqref{eq2}, detailed in \ref{square_operation}-\ref{eqsimilarity}.

\begin{equation} \label{eq2} 
	\mathcal{N}(\mathbf{z}_{i}) = \log\left(1 + \max_{\tilde{\mathbf{z}} \in \text{patches}(\mathbf{z}_{i})} \frac{1}{\|\tilde{\mathbf{z}} - \mathcal{P}_{i}\|_{2}^{2} + \epsilon}\right)
\end{equation}

Assume that the prototype vector has dimensions of \( [1, 1, k] \), where \( k \) is less than the length \( L' \) of the time series. The time patch vector \( x \) has dimensions \( [B, C, L'] \). The detailed steps to compute the L2 distance are as follows:
	
	\begin{enumerate}
	
	\item  Square the input time patch features $x$ as Eq. \ref{square_operation}.
	\begin{equation}
		x^2_{i,j} = (x_{i,j})^2
		\label{square_operation}
	\end{equation}
	
	\item Use convolution to compute the sum of squares within a local region as Eq. \ref{eqconvolution_sum_squares}. In Eq. \ref{eqconvolution_sum_squares}, \( k \) is the length of the convolution kernel.
	\(x_{i,j+k}\)~is element of the input feature \(x\) at position \((i, j+k)\).
	\(k\) is length of the \(\text{proto\_vector}\).
	\begin{equation}
		\text{x2\_patch\_sum}_{i,j} = \sum_{k=0}^{H-1} x_{i,j+k}^2
		\label{eqconvolution_sum_squares}
	\end{equation}

	\item  Compute the sum of squares of the prototype vector as Eq. \ref{eqsum_squares_proto_vector}. In Eq. \ref{eqsum_squares_proto_vector},
	\(\text{proto\_vector}_{k}\) is element of the prototype vector at position \(k\).
	
	\begin{equation}
		p2 = \sum_{k=0}^{H-1} (\text{proto\_vector}_{k})^2
		\label{eqsum_squares_proto_vector}
	\end{equation}

	\item  Compute convolution similarity,
	use convolution to compute the dot product between the feature vector and the prototype vector as Eq. \ref{eqconvolution_similarity}. 
	
	\begin{equation}
		\text{xp}_{i,j} = \sum_{k=0}^{L'-1} x_{i,j+k} \cdot \text{proto\_vector}_{k}
		\label{eqconvolution_similarity}
	\end{equation}

\item Compute the L2 distance,
	combine the sum of squares and the dot product to get the final L2 distance as Eq. \ref{l2_distance}. In Eq. \ref{l2_distance}, 	 \(\text{x2\_patch\_sum}_{i,j}\) is sum of squared values of the input feature \(x\) over a local patch.
	\(p2\) is sum of squared values of the \(\text{proto\_vector}\).
	\(\text{xp}_{i,j}\) is dot product of the input feature \(x\) and the \(\text{proto\_vector}\).
	\begin{equation}
		\text{distances}_{i,j} = \sqrt{\max \left(0, \text{x2\_patch\_sum}_{i,j} + p2 - 2 \cdot \text{xp}_{i,j} \right)}
		\label{l2_distance}
	\end{equation}

			\item 
		Compute similarity as Eq. \ref{eqsimilarity}.
	\begin{equation}
		\text{similarity}_{i,j} = \log \left( \max\left(1 + \frac{ 1}{\text{distance}_{i,j} + 1 \times \epsilon}\right) \right)
		\label{eqsimilarity}
	\end{equation}
	
\end{enumerate}

\color{black}

\item
\textbf{ Routing Score} \label{secrs}

\begin{flushleft}

	The routing score $\mathcal{N(\mathbf{z_i})}$ of the $i$-th tree node represents the similarity between the nearest reshaped time patch $\mathbf{\tilde{z}} $ 
	and the prototype $\mathcal{P}_i$. The routing score determines the branching direction (left or right), as shown in Fig.\ref{overview}(b). We define the routing score for each child node as briefly described in Eq.\eqref{eq3}, detailed in Eq. \ref{maxim}-\ref{final_dists}.

	\begin{equation} \label{eq3} 
		\mathcal{R}_{i, j}\left(\mathbf{z}_i\right) =
		\begin{cases}
			\left[\mathcal{N}\left(\mathbf{z}_i\right)\right]_0^1 & \text{if } j = 2 \times i \\
			1 - \left[\mathcal{N}\left(\mathbf{z}_i\right)\right]_0^1 & \text{if } j = 2 \times i + 1
		\end{cases}
	\end{equation}
	
\end{flushleft}

	\begin{enumerate}

		\item 
	In Eq. \ref{maxim},
		\(\text{similarity}\) is similarity scores computed from the distance.
		\begin{equation}
			\text{maxim} = \text{maxpool1d}(\text{similarity})
			\label{maxim}
		\end{equation}

		\item 
		In Eq. \ref{eq:to_left} nd \ref{eq:to_right}, \(\text{to\_left}\) is probability of choosing the left branch, to\_left takes the values from the 1st column of all rows.
		\(\text{maxim}\) is maximum similarity values after maximum pooling.
		\begin{equation}
			\text{to\_left} = \text{maxim}_{:, 0}
			\label{eq:to_left}
		\end{equation}
	
		\begin{equation}
			\text{to\_right} = 1 - \text{to\_left}
			\label{eq:to_right}
		\end{equation}

\item
In Eq. \ref{eqldis} and \ref{eqrdis}, \(\text{logits}\) is prediction from ST model.
attention\_l, attention\_r are functions that applies the attention model (Section \ref{att}) on the input patches for the left branch, enhancing relevant features.
	\begin{equation} \label{eqldis}
		\text{l\_dists} = \text{left}(\text{logits},~ \text{attention\_l}(\text{patches}))
	\end{equation}
	\begin{equation} \label{eqrdis}
		\text{r\_dists} = \text{left}(\text{logits},~ \text{attention\_r}(\text{patches}) )
	\end{equation}

	\item 
	In Eq. \ref{final_dists},  \(\text{dists}\) is the computed distance values in current node, combining distance information from both the left and right branches.
	\(\text{to\_left}\) is the probability of choosing the left branch, calculated from the maximum similarity values. The value typically ranges between [0, 1].
	\(\text{to\_right}\) is the probability of choosing the right branch, computed as \(1 - \text{to\_left}\). 
	\(\text{l\_dists}\) (r\_dists) is distance from the left (right) branch, indicating the distance between the feature vector and the prototype vector of the left (right) branch.
	\begin{equation}
			\text{dists} = \text{to\_left} \times \text{l\_dists} + \text{to\_right} \times \text{r\_dists}
			\label{final_dists}
	\end{equation}

	
	\end{enumerate}

\end{enumerate}

\color{black}

\subsubsection{Attention module} \label{att}
We propose an attention module that utilizes spatial and channel attention mechanisms for feature refinement and extraction. We introduce the context converter module to capture more distinct time patches. Finally, the context time patch $\mathbf{z_i}$ enhanced by attention module ($V_{i,j}$) to get $\mathbf{z_j}$ using Eq.~\eqref{eq6}, and next $\mathbf{z_j}$ will be fed into the corresponding child nodes. Assume that the dimension of patch 
 is $B \times C \times L'$. The structure of  attention module is shown in Fig.~\ref{overview} (c).

\textbf{Channel Attention}: the spatial information of the feature graph is first aggregated using average and maximum pooling operations to generate two different spatial context descriptors representing the average and maximum pooling features, respectively. Finally, the two are spliced. The channel attention is computed as Eq.~\eqref{eq4}. In Eq.~\eqref{eq4}, the dimension of $p$ is $B \times C \times 1$.

\textbf{Spatial Attention}: the relationship between feature spaces is utilized to generate a spatial attention map, which is complementary to channel attention. The average pooling operation and the maximum pooling operation are first applied along the channel axis generate a valid feature descriptor. The spatial attention is calculated as Eq.~\eqref{eq5}. In Eq.~\eqref{eq5}, the dimension of $p_j$ is $B \times 1 \times L'$. In Eq.~\eqref{eq4} and Eq.~\eqref{eq5}, $c^a$ is one-dimensional convolution operation with filter size $a$.

\begin{equation} \label{eq4}
	q = \operatorname{sigmoid}\left(c^a\left[\operatorname{avgpool1d}\left(q_i\right) ; \operatorname{maxpool1d}\left(q_i\right)\right]\right) 
\end{equation}

\begin{equation}\label{eq5}
	q_j = \operatorname{sigmoid}\left(c^a\left[\operatorname{avgpool1d}(q) ; \operatorname{maxpool1d}(q)\right]\right) 
\end{equation}

\begin{equation}\label{eq6}
	\mathbf{z}_j = V_{i,j}\left(\mathbf{z}_i\right) = \operatorname{attention}\left(\mathbf{z}_i\right) 
\end{equation}

\subsubsection{Prediction module}

\paragraph{\textbf{Routing score}}
The leaf node in each tree module corresponds to the leaf prediction module $\mathcal{L}(\cdot)$ used to predict the category probabilities. Let $\mathbf{\rho(z_0)}$ be the cumulative routing score of the output $\mathbf{z_0}$ from the root node to the $l$-th leaf node on a set of edges of a particular path $p_l$. The cumulative routing score $\mathbf{\rho(z_i)}$ is computed by using Eq.~\eqref{eqpred}. In Eq. \ref{eqpred}, \(\mathcal{R}_{i,j}\) indicates the routing from node \(i\) to child node \(j\), detailed in Eq. \ref{eq3}. 
\(\mathcal{E}_{i,j}\left(\mathbf{z}_i\right)\) represents the function that transforms the input feature \(\mathbf{z}_i\) for node \(i\) and child node \(j\).
\(\text{children}(i)\) denotes the set of all child nodes of node \(i\).

\begin{equation}\label{eqpred}
\rho\left(\mathbf{z}_i\right) = \text{Aggregate}\left(\left\{ \mathcal{R}_{i,j}\left(\mathcal{E}_{i,j}\left(\mathbf{z}_i\right)\right) \mid j \in \text{children}(i) \right\}\right)
\end{equation}

\paragraph{\textbf{FC}}
Each leaf prediction module passes through the time patch fully connected layer (timeFC), and tree fully connected layer (treeFC), as shown in Fig.~\ref{overview} (d). The leaf prediction module is computed as Eq.~\eqref{eq8}. In Eq.~\eqref{eq8}, treeFC is obtained based on a certain path, and timeFC is obtained through time patch module. Thus, the module can combine local path and global contextual feature representations. The detailed steps are shown as Eq. \ref{eqx} -  \ref{eqdistsfinal}.

\begin{equation}\label{eq8}
	\begin{aligned}
		&\begin{gathered}
			\mathcal{L}\left(\mathbf{z}_l, x\right)=\operatorname{treeFC}\left(\mathbf{z}_l\right)+\operatorname{timeFC}(\mathbf{z}_0)
		\end{gathered}
	\end{aligned}
\end{equation}

	\begin{enumerate}
	
	\item
	Use max pooling to process the features of the nodes in the layer preceding the leaf as Eq. \ref{eqx}. In Eq. \ref{eqx}, $\text{patches}''$ is the feature from the layer preceding the leaf.
	
	\begin{equation}
		x = \text{maxpool1d}(\text{patches}'') 
		\label{eqx}
	\end{equation}

	\item
	Obtain the final prediction through path prediction and ST prediction  as Eq. \ref{eqx}.   In Eq. \ref{eqx}, \(\text{pred}(x)\) are predictions obtained from a MLP layer in the leaf node, logits are predictions of ST module.
	
	\begin{equation}
		\text{dists}' = \text{pred}(x) + \text{logits}
		\label{eqdists}
	\end{equation}

	\item
	Use softmax function to obtain the final prediction  as Eq. \ref{eqx}. In \ref{eqx}, \(\text{pred}(x)\) are predictions obtained from a MLP layer in the leaf node, logits are predictions of ST module.
	\begin{equation}
		\text{dists\_leaf} = \text{softmax}(\text{dists}'
		)
		\label{eqdistsfinal}
	\end{equation}

\end{enumerate}

\color{black}
\paragraph{\textbf{Loss}}

The ultimate prediction $\hat{y}$ is obtained through the multiplication of all leaf predictions $\mathcal{L}$ with the sum of the accumulated routing scores $\rho_l$, as illustrated in Eq.~\eqref{eqloss1}-Eq.~\eqref{eqloss4}. The optimization of $\hat{y}$ is achieved by utilizing the Cross-Entropy loss function with the true label $y$.
The final prediction $\hat{y}$ is computed by multiplying all leaf predictions $\mathcal{L}$ by the sum of the accumulated routing scores $\rho_l$, as shown in Eq.~\eqref{eqloss1}-Eq.~\eqref{eqloss4}.

\begin{equation} \label{eqloss1}
	\hat{y} = \mathbf{\rho}^T \cdot \mathbf{g}(x)
\end{equation}
\begin{equation}
	g_l(x) = \sigma\left(\mathcal{L}(x, \mathbf{z}_l)\right)
\end{equation}

\begin{equation}
	\mathbf{\rho} = [\rho_1(\mathbf{z}_1), \rho_2(\mathbf{z}_2), \dots, \rho_n(\mathbf{z}_n)]
\end{equation}

\begin{equation}\label{eqloss4}
	\mathbf{g}(x) = [g_1(x), g_2(x), \dots, g_n(x)]
\end{equation}

In Eq. \ref{eqloss1},
	$\mathbf{\rho}$ is a vector of routing scores  for each node $l$, representing the similarity between each time patch $\mathbf{z}_l$ and its prototype.	
	 $\mathbf{g}(x)$ is a vector of functions $g_l(x)$ that apply an activation function $\sigma(\cdot)$ to the leaf $\mathcal{L}(x, \mathbf{z}_l)$ for each node.

The loss function for leaf nodes to make predictions is shown in Eq.~\eqref{eq10}. In Eq.~\eqref{eq10}, $N$ is the training sample set, $M$ is the number of labels, $y$ is the true label, and $\hat{y}$ is the model prediction label.

\begin{equation}\label{eq10}
	\begin{aligned}
		&\begin{gathered}
			Loss=-\frac{1}{N} \sum_{i=1}^N \sum_{j=1}^M y_{i, j} \log p\left(\hat{y}_{i, j}\right)
		\end{gathered}
	\end{aligned}
\end{equation}

\section{Experiment}

\subsection{Experimental setting}

\subsubsection{Datasets}
\
\newline We used 10 multivariate benchmark datasets from the UEA archive \footnote{public datasets: \url{http://timeseriesclassification.com/dataset}} for our comparison experiments. These datasets are mainly categorized as: Motion, Electroencephalography/Magnetoencephalography (EEG/MEG), Human Activity Recognition (HAR), Audio Spectra (AS). More information of each dataset are shown in Table~\ref{tab1}. Here is the dataset description.

\begin{table}[]
	\centering
	\caption{Brief information regarding the 10 benchmark datasets.}
	\label{tab1}
	\footnotesize
	\begin{tabular}{lllllll}
		\hline
		Datasets & Type & Train & Test & Dimensions & Lengths & Classes \\ \hline
		AWR & Motion & 275 & 300 & 9 & 144 & 25 \\
		AF & ECG & 15 & 15 & 2 & 640 & 3 \\
		BM & HAR & 40 & 40 & 6 & 100 & 4 \\
		CT & Motion & 1422 & 1436 & 3 & 182 & 20 \\
		\begin{tabular}[c]{@{}l@{}}HMD\end{tabular} & EEG/MEG & 160 & 74 & 10 & 400 & 4 \\
		NATOPS & HAR & 180 & 180 & 24 & 51 & 6 \\
		PD & Motion & 7494 & 3498 & 2 & 8 & 10 \\
		SR2 & EEG/MEG & 200 & 180 & 7 & 1152 & 2 \\
		SAD & AS & 6599 & 2199 & 13 & 93 & 10 \\
		SWJ & ECG & 12 & 15 & 4 & 2500 & 3 \\ \hline
	\end{tabular}
\end{table}

ArticularyWordRecognition (AWR): The AWR dataset comprises 12 sensors providing X, Y, and Z time series positions at a sampling rate of 200 Hz.
AtrialFibrillation (AF): The AF dataset consists of two-channel ECG recordings aimed at predicting spontaneous termination of AF. Each instance in the dataset represents a 5-second segment of AF, containing two ECG signals sampled at a rate of 128 samples per second. 

BasicMotions (BM): The BM dataset was generated by four students performing four different activities while wearing a smartwatch. It includes four classes: walking, resting, running, and badminton. 

CharacterTrajectories (CT): The CT dataset was captured using a tablet and includes three dimensions: x-coordinate, y-coordinate, and pen tip force. The data was recorded at a sampling rate of 200 Hz. 

HandMovementDirection (HMD): The HMD dataset includes MEG activity modulated based on the direction of wrist movements. The data was recorded while subjects performed wrist movements in four different directions.

NATOPS: The NATOPS dataset is generated using sensors placed on the hands, elbows, wrists, and thumbs. It records the x, y, and z coordinates for each of the eight sensor locations.

PenDigits: The PenDigits dataset is used for handwritten digit classification. It involves 44 different writers drawing digits from 0 to 9. Each instance in the dataset consists of the x and y coordinates of the pen as it traces the digit on a digital screen.

SelfRegulationSCP2 (SR2): The SR2 dataset was obtained from an artificially respirated patient. The subject controlled the movement of a cursor on a computer screen while cortical potentials were recorded.

SpokenArabicDigits (SAD): The SAD dataset comprises 8800 time series instances representing spoken Arabic digits. Each instance is composed of 13 Frequency Cepstral Coefficients extracted from the speech signals.

StandWalkJump (SWJ): The SWJ dataset includes short-duration ECG signals recorded from a healthy 25-year-old male performing various physical activities. The dataset aims to study the effect of motion artifacts on ECG signals and their sparsity.

\subsubsection{Experimental Environment}
All models were trained in a Python 3.8 environment using PyTorch 1.10.0 with Cuda 11.3. The training environment was configured with the Ubuntu 20.04 operating system, equipped with an NVIDIA GeForce RTX 2080 GPU (20GB VRAM) and an Intel(R) Xeon(R) Platinum 8352V CPU (48GB RAM).

\subsubsection{Experimental Parameters}
 For all experiments in this paper, ST-Tree utilized a three-layer neural tree module and a one-layer time patch module. The model was trained using the Adam optimizer and a Cross-Entropy loss function. The results reported in this study were obtained after 50 epoches on all datasets.
To address model performance sensitivity to hyperparameters:
\begin{enumerate}
\item \textbf{Training Parameters.}
When using the Adam optimizer to update model parameters, the initial learning rate is set to 0.001, which controls the step size of weight updates. The learning rate decay period is set to 5, meaning that the learning rate will decrease if the loss does not improve after every 5 training epochs. The decay factor is set to 0.90, indicating that the learning rate will be reduced to 90\% of its original value after every update. This exponential decay strategy improves the model's convergence and stability by gradually reducing the learning rate, allowing for finer adjustments as the model approaches the optimal solution.
The batch size must fit within our GPU memory.  
Through testing, we found 50 epochs are sufficient for the model to converge.
\item \textbf{Model Parameters }
The ST-Tree employs a three-layer neural tree module and a one-layer time patch module.  
For the decision tree model, the depth affects the model's complexity and performance. We have discussed the impact of tree depth on model accuracy in Section \ref{sectreedepth}.  
For the time patch module, we use only one layer because the time patch relies on self-attention calculations, which have significant time and space complexity. Using multiple layers could lead to overfitting and difficulties in convergence, while one layer has already shown sufficient performance, as indicated in column ST-Tree (without tree) of Table \ref{tab3}. Therefore, for training convenience and performance, we use just one layer.
\end{enumerate}

\subsubsection{Reproducibility}
\
\newline For the purpose of reproducibility, we have made our codes and parameters available on GitHub\footnote{\href{https://github.com/dumingsen/ST-Tree}{https://github.com/dumingsen/ST-Tree}
	}. This allows for independent replication of the experimental results.

\subsection{Experimental performance}
\subsubsection{Experimental Results}
\
\newline To evaluate the performance of ST-Tree, we fairly conducted a comparison with various representative methods in the field of multivariate time series classification, including various traditional and deep learning based methods:
  DTW-1NN (\cite{Chen2013DTWDTS}),
   MLSTM-FCN (\cite{Karim2018MultivariateLF}),
     MTSC\_FF (\cite{Du2024MultivariateTS}),
     TST (supervised version) (\cite{Zerveas2020ATF}),
      ROCKET (\cite{Dempster2019ROCKETEF}),
      ShapeNet (\cite{Li2021ShapeNetAS}), 
     SRL (\cite{Franceschi2019UnsupervisedSR}),
      MiniRocket \cite{Dempster2019ROCKETEF}, 
      TStampTransformer (unsupervised version) \cite{Zerveas2020ATF}, TS-TCC (\cite{Eldele2021TimeSeriesRL}), 
      and TNC (\cite{Tonekaboni2021UnsupervisedRL}), totaling 9 methods. The accuracies of these methods, along with their corresponding parameters from publicly available data in their respective studies, are presented in Table~\ref{tab2}. The following is a brief description of the methods involved in the comparison.


\textbf{1NN-DTW}: Introduces a variation of the semi-supervised DTW algorithm.

\textbf{MLSTM-FCN}: Combines the LSTM, FCN, and Attention blocks to enhance accuracy in MTSC by incorporating a squeeze-and-excitation block into the FCN block.





\textbf{ShapeNet}: Projects MTS subsequences into a unified space and applies clustering to identify shapelets.

\textbf{Rocket, MiniRocket}: Utilize random convolutional kernels to extract features from univariate time series, with extensions made for MTSC.

\textbf{TNC}: Exploits the local smoothness of a signal to define temporal neighborhoods and learn generalizable representations.

\textbf{TS-TCC}: Promotes consistency among different data augmentations to learn transformation-invariant representations.

\textbf{MTSC\_FF}: Extracts fusion features by leveraging Transformer, GNN, and Attention mechanism modules to capture time domain spatial and frequency domain features, respectively.

\textbf{TST(S)}, \textbf{TST(U)}: Processes the values at each timestamp as input for a transformer encoder, Incorporates unsupervised and unsupervised pre-training scheme within the framework relatively.

\textbf{SRL}: Utilizes negative sampling techniques with an encoder-based architecture to learn representations.

\begin{figure}
	\centering
	\includegraphics[width=0.6\textwidth]{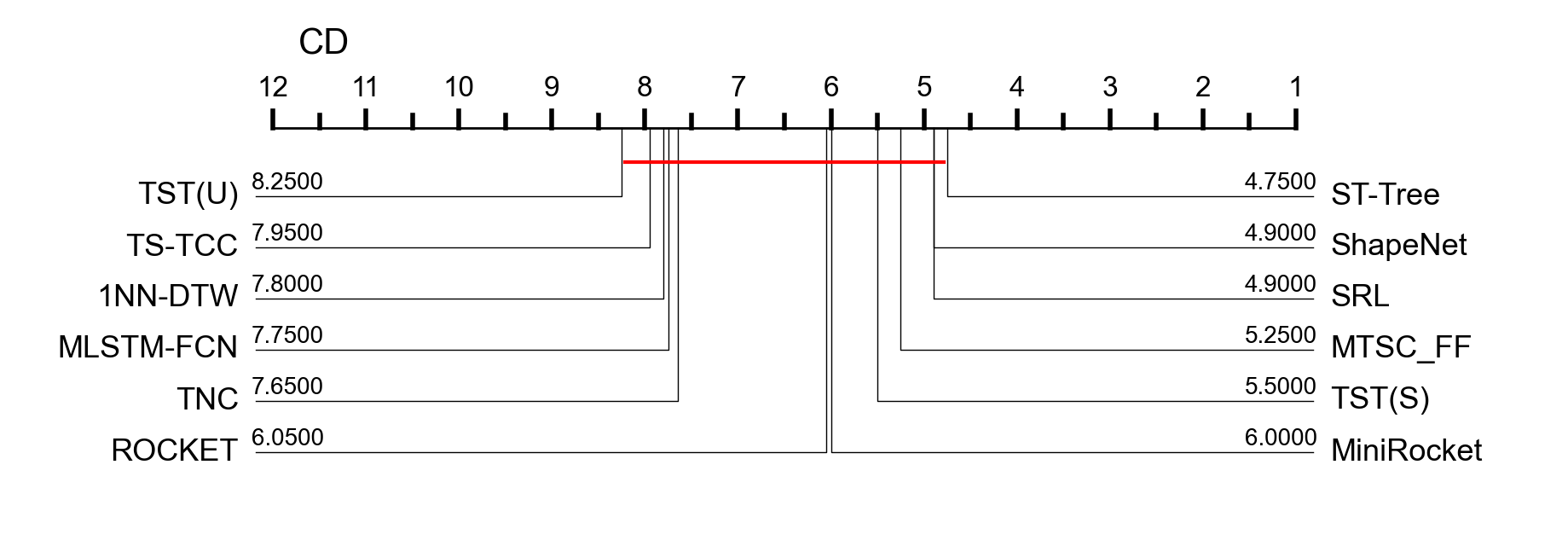}
	\caption{CD diagram of 9 implementations on 10 UEA datasets. }
	\label{cd}
\end{figure}

Based on the results provided in Table~\ref{tab2}, we clearly observe that MTSC\_FF has the highest average accuracy (0.789 ) on the 10 datasets. To visualize the comparison results more, we show the critical difference figure (CD) based on the accuracies in Table~\ref{tab2}. The CD figure ranks the 12 MTS classification classifiers in ascending order. As shown in Fig.~\ref{cd}, ST-Tree has the smallest rank. The CD figure similarly illustrates that our method is in the first tier.

ST-Tree exceeds the other methods in the dataset StandWalkJump, which, based on the characteristics of this dataset, indicates that ST-Tree is more capable of modeling long sequences. Among them, MTSC\_FF, MF-Net use self-attention and graph neural network to mine global, spatial and other features. ST-Tree is different from MF-Net and MTSC\_FF in terms of methodology and feature mining, the latter focuses on modeling global and spatial features, which is suitable for datasets with high dimensionality. ShapeNet projects the time series subsequence into a uniform space and applies clustering to find shape features. 
ST-Tree differs from ShapeNet in terms of methodology and feature extraction, as ST-Tree is a tree-based approach while ShapeNet uses clustering and shapelets features. Shapelets are precise and intuitive, while ST-Tree can visualize the decision process while being interpretable. TST takes the value of each timestamp as an input to the Transformer encoder. ST-Tree is a method based on ST combined with a tree structure, whereas TStamp Transformer uses the Transformer encoder and is not able to visualize the decision process.

\begin{table}[]
		\centering
	\caption{Experimental results from traditional and deep-learning implementations. Avg represents the average accuracy achieved by each method across 10 datasets and Win indicates the number of datasets for which a method achieved the best accuracy. The highest accuracy for each dataset is marked in bold.}
	\label{tab2}
	
\resizebox{1.1\textwidth}{!}{
		\begin{tabular}{lllllllllllll}
			\hline
			Dataset & \begin{tabular}[c]{@{}l@{}}MTSC\\ \_FF\end{tabular} & TST(S) & ROCKET & \begin{tabular}[c]{@{}l@{}}Shape\\ Net\end{tabular} & SRL & \begin{tabular}[c]{@{}l@{}}Mini\\ Rocket\end{tabular} & TST(U) & \begin{tabular}[c]{@{}l@{}}TSC\\ -TCC\end{tabular} & TNC & \begin{tabular}[c]{@{}l@{}}MLSTM\\ -FCN\end{tabular} & \begin{tabular}[c]{@{}l@{}}1NN\\ -DTW\end{tabular} & \begin{tabular}[c]{@{}l@{}}ST\\ -tree\end{tabular} \\ \hline
			AWR & 0.983 & 0.977 & \textbf{0.996} & 0.987 & 0.987 & 0.992 & 0.983 & 0.953 & 0.973 & 0.973 & 0.980 & 0.980 \\
			AF & \textbf{0.533} & 0.067 & 0.249 & 0.400 & 0.133 & 0.133 & 0.200 & 0.267 & 0.133 & 0.267 & 0.267 & 0.466 \\
			BM & 0.950 & 0.975 & 0.990 & \textbf{1.000} & \textbf{1.000} & \textbf{1.000} & 0.975 & \textbf{1.000} & 0.975 & 0.950 & \textbf{1.000} & 0.950 \\
			CT & 0.986 & 0.975 & 0.967 & 0.980 & \textbf{0.994} & 0.993 & N/A & 0.985 & 0.967 & 0.985 & 0.969 & 0.970 \\
			HMD & 0.541 & 0.243 & 0.446 & 0.338 & 0.270 & 0.392 & \textbf{0.608} & 0.243 & 0.324 & 0.365 & 0.306 & 0.554 \\
			NATOPS & 0.889 & 0.850 & 0.885 & 0.883 & \textbf{0.944} & 0.928 & 0.900 & 0.822 & 0.911 & 0.889 & 0.850 & 0.916 \\
			PD & 0.979 & 0.560 & \textbf{0.996} & 0.977 & 0.983 & N/A & 0.974 & 0.974 & 0.979 & 0.978 & 0.939 & 0.980 \\
			SR2 & 0.511 & \textbf{0.589} & 0.533 & 0.578 & 0.556 & 0.522 & \textbf{0.589} & 0.533 & 0.550 & 0.472 & 0.533 & 0.550 \\
			SAD & 0.990 & \textbf{0.993} & 0.712 & 0.975 & 0.956 & 0.620 & \textbf{0.993} & 0.970 & 0.934 & 0.990 & 0.960 & 0.990 \\
			SWJ & 0.467 & 0.267 & 0.456 & \textbf{0.533} & 0.400 & 0.333 & 0.267 & 0.333 & 0.400 & 0.067 & 0.333 & \textbf{0.533} \\
			Avg & 0.783 & 0.650 & 0.723 & 0.765 & 0.722 & 0.657 & 0.721 & 0.708 & 0.715 & 0.694 & 0.714 & \textbf{0.789} \\
			Win & 1 & 2 & 2 & 1 & \textbf{3} & 1 & \textbf{3} & 1 & 0 & 0 & 1 & 1 \\ \hline
		\end{tabular}
	}
	\end{table}

\subsection{Ablation analysis}

\begin{figure}
	\centering
	\subfloat[ST-Tree (without tree)]{
		\centering
		\includegraphics[width=0.4\textwidth]{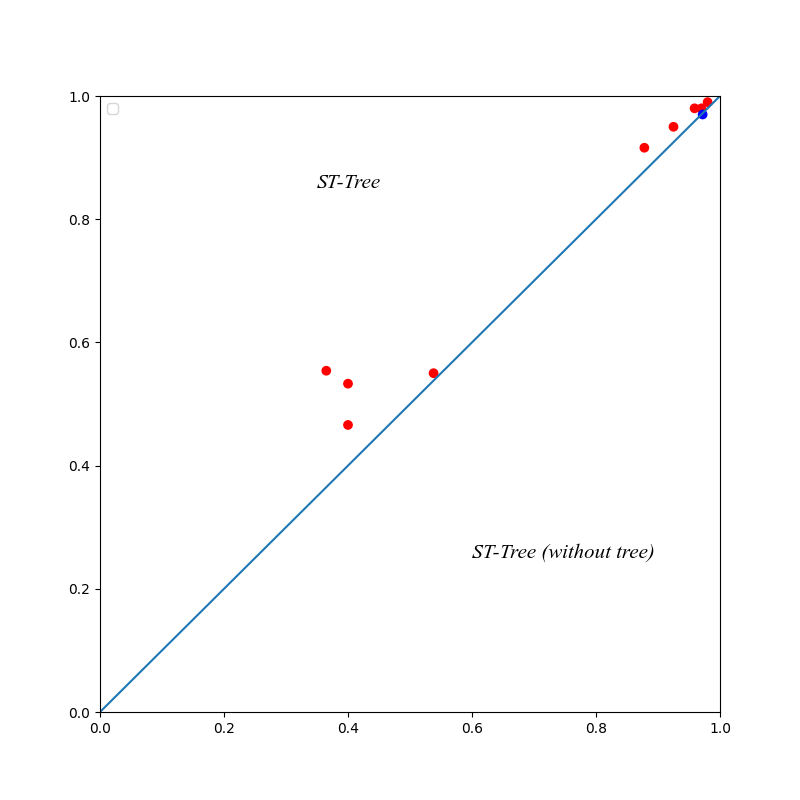}
		\label{ab1}
	}
	\subfloat[ST-Tree (without attention)]{
		\centering
		\includegraphics[width=0.4\textwidth]{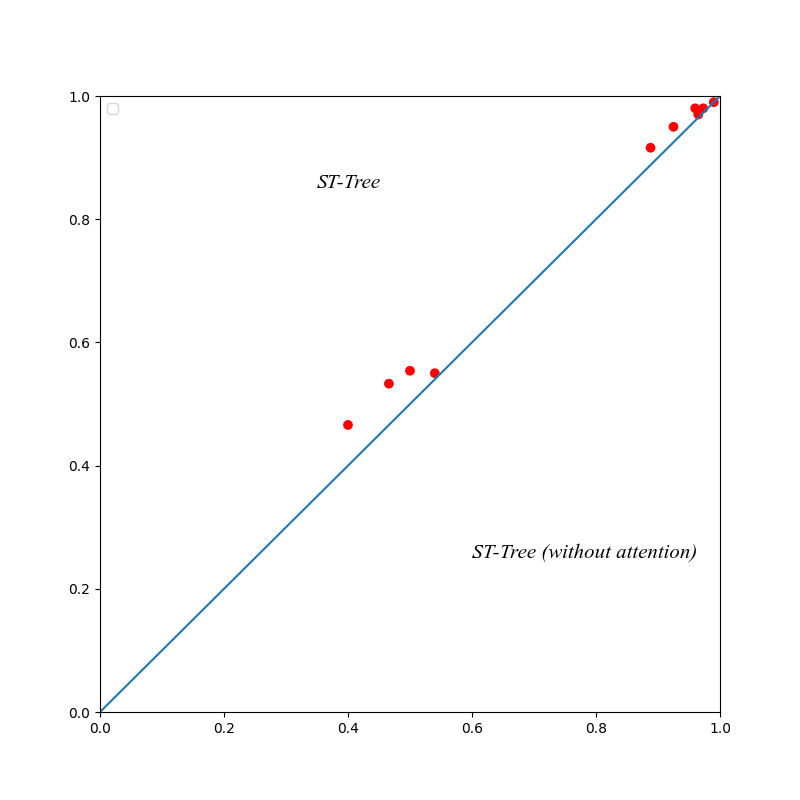}
		\label{ab2}
	}
	
	\caption{Comparison of the accuracy of ST-Tree and ST-Tree (without tree), ST-Tree (without attention), the closer to the upper left corner ST-Tree works better.}
	\label{ab}
\end{figure}

\begin{table}[]
	\centering
	\caption{Ablation study about neural tree module and attention module.}
	\label{tab3}
	\footnotesize
	\begin{tabular}{llll}
		\hline
		Datasets & ST-Tree & \begin{tabular}[c]{@{}l@{}}ST-Tree \\ (without tree)\end{tabular} & \begin{tabular}[c]{@{}l@{}}ST-Tree \\ (without attention)\end{tabular} \\  \hline
		AWR & \textbf{0.980} & 0.970 & 0.973 \\
		AF & \textbf{0.466} & 0.400 & 0.400 \\
		BM & \textbf{0.950} & 0.925 & 0.925 \\
		CT & 0.970 & \textbf{0.972} & 0.965 \\
		HMD & \textbf{0.554} & 0.365 & 0.500 \\
		NATOPS & \textbf{0.916} & 0.878 & 0.888 \\
		PD & \textbf{0.980} & 0.959 & 0.960 \\
		SR2 & \textbf{0.550} & 0.538 & 0.540 \\
		SAD & \textbf{0.990} & 0.980 & \textbf{0.990} \\
		SWJ & \textbf{0.533} & 0.400 & 0.466 \\
		AVG & 0.789 & 0.739 & 0.761 \\ 
		Win & \textbf{9} & 1 & 1 \\  \hline
	\end{tabular}
\end{table}
The neural tree module and the attention module were removed for validating the effectiveness of each module, respectively. ST-Tree obtained the best "Win" (9) and the highest average accuracy (0.789). As shown in Fig.~\ref{ab}, the experimental results illustrate the positive effects of the neural Tree module and the attention module for ST-Tree.

\subsection{Feature interpretability analysis}

\begin{figure}
	\centering
	\begin{minipage}[t]{0.5\linewidth}
		\centering
		\includegraphics[height=4cm,width=4cm]{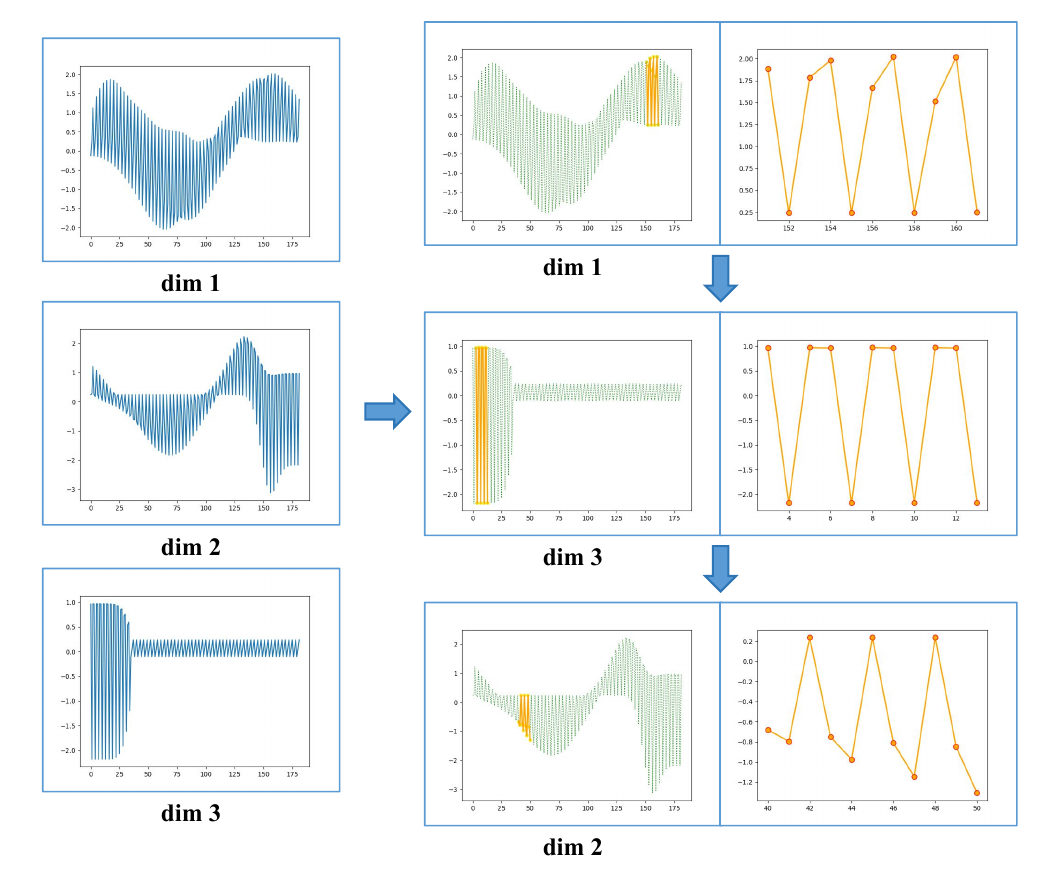}
		\caption{CharacterTrajectories.}
		\label{treeCha}
	\end{minipage}%
	\begin{minipage}[t]{0.5\linewidth}
		\centering
		\includegraphics[height=4cm,width=4cm,]{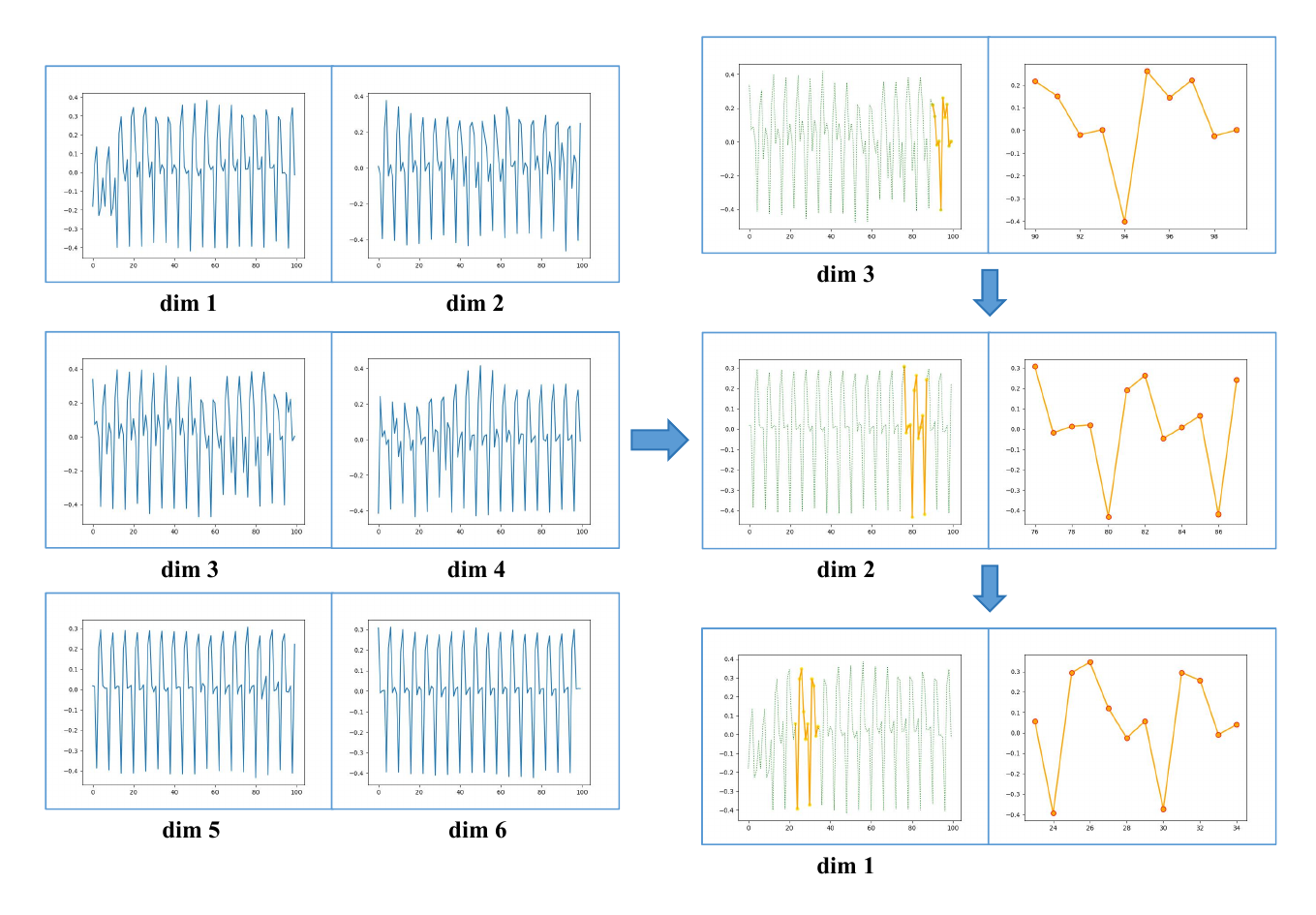}
		\caption{BasicMotions.}
		\label{treebasic}
	\end{minipage}
\end{figure}

\begin{figure}
	\centering
	\includegraphics[width=0.5\textwidth]{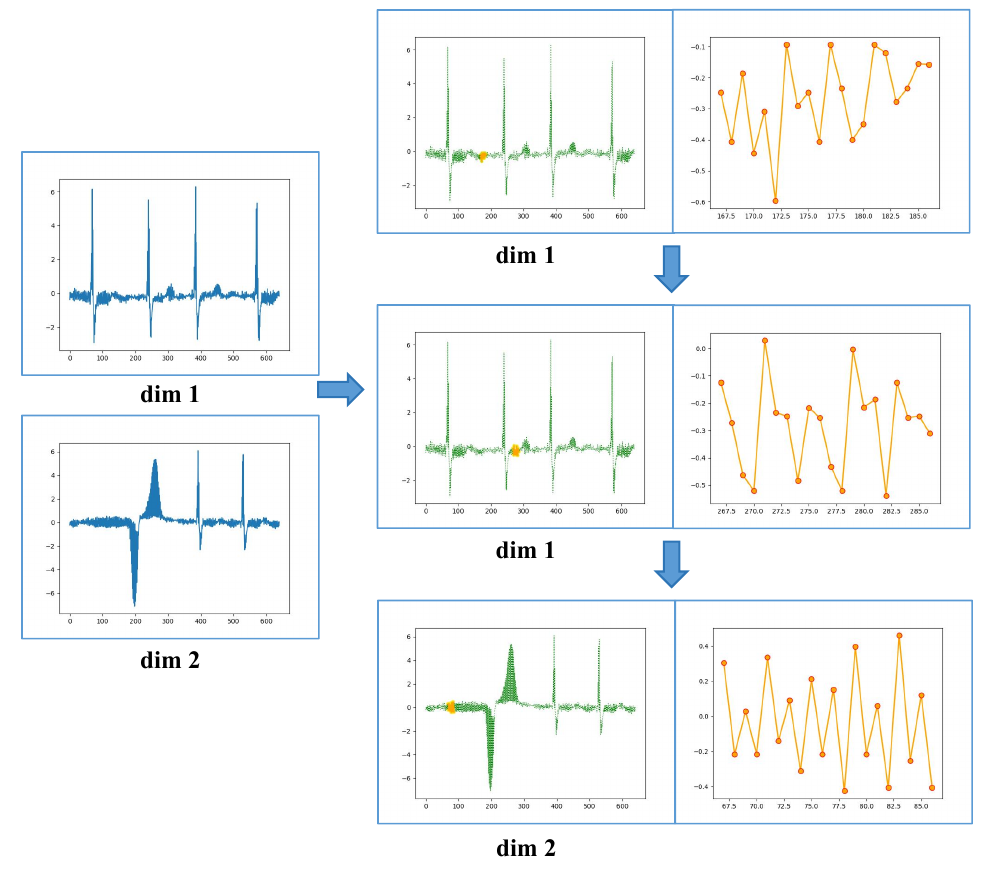}
	\caption{AtrialFibrillation.}
	\label{treeAtr}
\end{figure}

\begin{figure}
	\centering
	\includegraphics[width=0.5\textwidth]{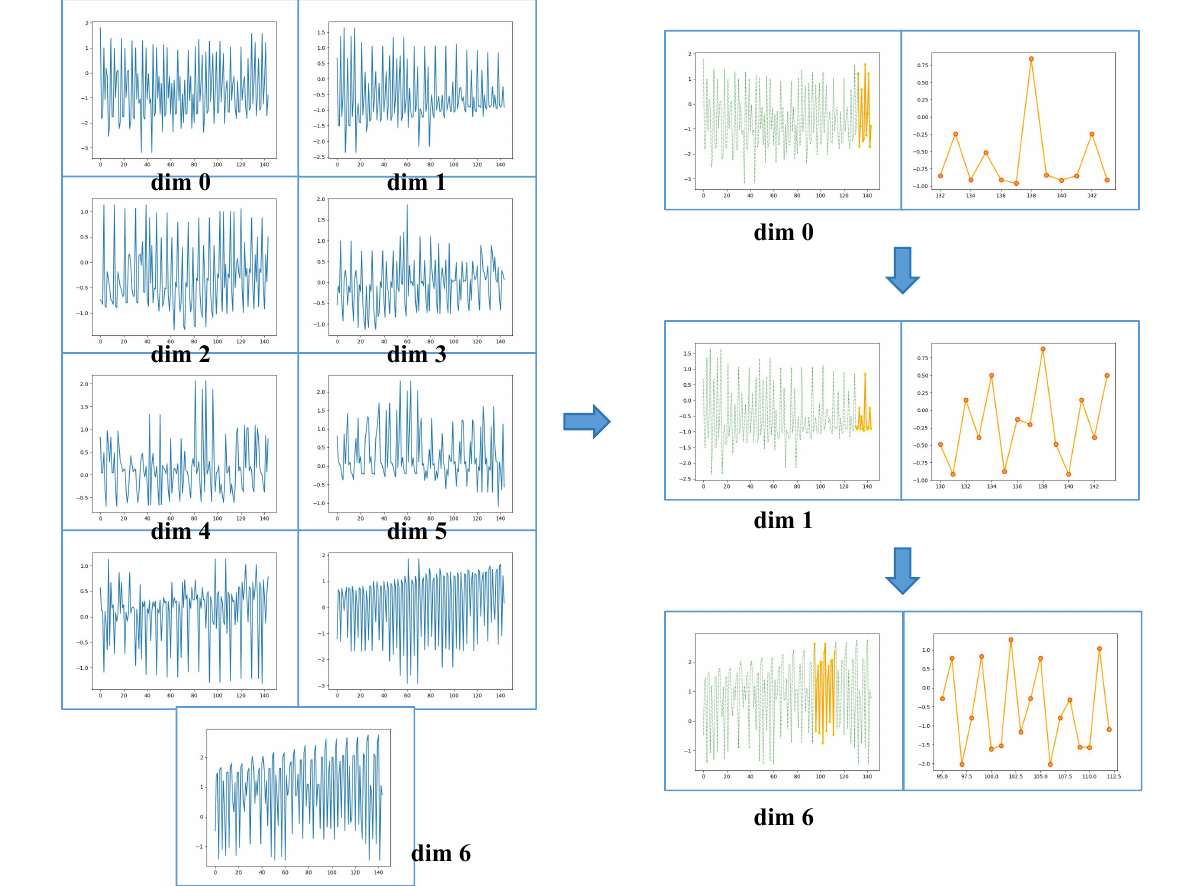}
	\caption{ArticularyWordRecognition.}
	\label{treeAWR}
\end{figure}

We visualized the decision-making process during classification on the BasicMotions, CharacterTrajectories and AtrialFibrillation datasets, as shown in Fig.~\ref{treeCha}, \ref{treebasic}, \ref{treeAtr} and \ref{treeAWR}.. The orange parts of the figure are the prototype bases corresponding to the tree nodes, and the arrows represent the decision order. 

\subsection{Tree depth analysis} \label{sectreedepth}
\begin{figure}
	\centering
	\includegraphics[width=0.6\textwidth]{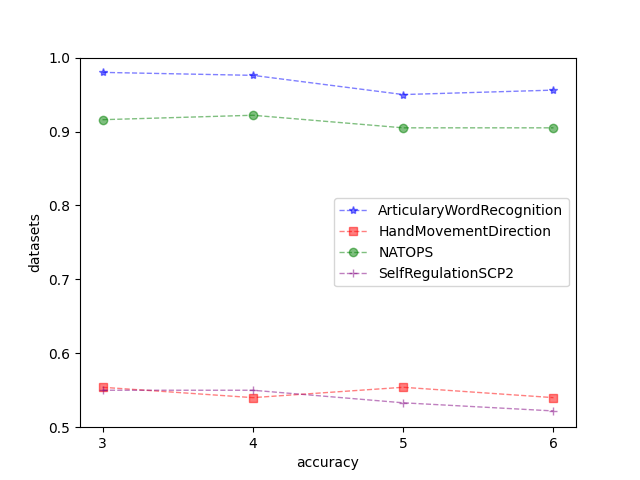}
	\caption{ Variation in the precision of the number of layers of the corresponding tree for different datasets.}
	\label{treedepth}
\end{figure}

We analyzed the model and tree depth against the ArticularyWordRecognition, HandMovementDirection, NATOPS and SelfRegulationSCP2 datasets. The experimental results show that we can see that the tree depth of the ST-Tree varies with the complexity of the dataset and that increasing the depth does not improve the performance but leads to overfitting. According to Fig.~\ref{treedepth}, we can analyze the effect of the number of layers of the tree corresponding to different datasets on the accuracy.

For Articulatory Word Recognition, as the number of layers in the tree increases from 3 to 6, accuracy decreases. Therefore, increasing the tree's depth results in reduced accuracy, making a tree depth of 4 layers preferable.
For HandMovementDirection, increasing the number of layers in the tree does not enhance accuracy; instead, the accuracy remains stable regardless of the number of layers. This may indicate that the number of tree layers does not have a significant effect on the accuracy in this particular dataset. For NATOPS, the accuracy improved when the number of layers in the tree was increased from 3 to 4. This shows that increasing the number of layers in the tree can improve the accuracy, but further increasing the number of layers may not have a significant effect. Therefore, it is better to keep the tree at 4 layers. For SelfRegulationSCP2, the increase in the number of layers of the tree leads to a continuous decrease, so a 3-layer tree works best. In summary, when choosing the number of layers in the tree, the specific dataset characteristics and performance requirements need to be considered.

\subsection{Robustness to Noise}

Our method combats noise in the following ways.
\begin{enumerate}
	\item \textbf{Data preprocessing}. We implement data preprocessing steps, such as normalization and standardization. This processing enhances the overall quality of the data, making the model more effective during training.
	\item \textbf{Self-attention mechanism}. By using the self-attention mechanism, the model can automatically focus on the most relevant parts of the input sequence, reducing the impact of noise on the decision-making process while learning good representations of the time series. This mechanism allows the model to extract useful features even in noisy environments.
	\item \textbf{Convolution for similarity calculation}. The convolution layer computes the sum of squares of input features, which amplifies the characteristics of the true signal while reducing noise interference. Combined with the ReLU activation function, it effectively removes negative values, further enhancing the robustness of the model.
\end{enumerate}

\subsection{Transfer Learning}

To reduce training and computational costs, we propose the following strategies, the detailed process is shown in Algorithm \ref{algtransfer}.
\begin{enumerate}
	\item \textbf{Reduce the number of epochs}. Monitor the performance on the validation set during training. If there is no significant improvement over several training epochs, stop training early.
	\item \textbf{Learning rate scheduling}. By dynamically adjusting the learning rate, we can converge to the optimal solution more quickly.
	\item \textbf{Transfer learning}. In the phase of obtaining time series representations, leverage the knowledge of pre-trained models by fine-tuning them to adapt to new tasks.
\end{enumerate}
\begin{algorithm} 	
	\caption{Training Process for ST-Tree with Enhanced Strategies and Transfer Learning}
	\label{algtransfer}
	\footnotesize
	\begin{algorithmic}[1]
		\Require Training set $\mathcal{T}$, number of epochs $E$, ST module $f$, ST parameters $\theta$, tree model parameters $\omega$, validation set $\mathcal{V}$, patience $P$, initial learning rate $\alpha$, pre-trained parameters $\theta_{pre}$
		, $decay\_steps=5, decay\_rate = 0.90$
		
		\State Initialize $learning\_rate$
		\State Initialize best validation performance $best\_val\_perf$ and counter $patience\_counter = 0$
		
		\State Load pre-trained parameters $\theta \gets \theta_{pre}$ 
		\Comment{Transfer learning initialization}
		\State Ignore the first layer of $\theta$  \Comment{Adapt to new dataset dimensions}
		
		\For{$e \in\{1, \ldots, E\}$}
		\State Split $\mathcal{T}$ into $B$ batches $\{\mathcal{T}_1, \ldots, \mathcal{T}_B\}$ \Comment{Batch splitting process}
		
		\For{batch $b \in \{1, \ldots, B\}$}
		\State Extract batch $\left(x_b, y_b\right)$ from $\mathcal{T}_b$ 
		
		\State $\mathbf{z}_0^b \gets f(x_b; \theta)$ \Comment{Fine-tuning}
		
		\State  $\hat{y}_b \gets \Call{traverse}{(1, \mathbf{z}_0^b; \omega)}$ \Comment{Traverse the tree to get predictions}
		
		\State  $\text{loss}(\hat{y}_b, y_b)$ \Comment{Calculate the loss based on predictions}
		
		\State Update $\theta$ and $\omega$  \Comment{Optimize model parameters using Adam optimizer}
		\EndFor
		
		\State Evaluate model on validation set $\mathcal{V}$ and obtain $val\_perf$
		
		\If{$val\_perf > best\_val\_perf$}
		\State $best\_val\_perf \gets val\_perf$
		\State $patience\_counter \gets 0$  \Comment{Reset patience counter}
		\Else
		\State $patience\_counter \gets patience\_counter + 1$
		\EndIf
		
		\If{$patience\_counter > P$} \Comment{Early stopping condition}
		\State \textbf{break} \Comment{Stop training if no improvement}
		\EndIf
		
		\State \quad $learning\_rate$ $\gets$$ learning\_rate$ $\times$ $decay\_rate^{(e / decay\_steps)} $
		
		\Comment {Update the optimizer with the new learning rate}
		
		\EndFor 
		
	\end{algorithmic}
\end{algorithm}

\section{Conclusion}
We proposed ST-Tree with Interpretability for multivariate time series classification, which consists of a time patch module and a neural tree module. specifically, the time patch module uses ST as a feature extractor and time patch generator. To improve the interpretability of the model and visualize the decision-making process, the generated time patches are fed into a tree model, the neural tree module, which hierarchically paths the multivariate time series through a binary tree. Each of these nodes corresponds to a trained time series training prototype and is used to compute the similarity with the time series during the classification process. Finally, the leaf nodes perform label prediction on the multivariate time series by calculating branch routing scores through the leaf prediction module. Experimental evaluation on 10 UEA datasets demonstrated that the ST-Tree model has improved accuracy on the multivariate time series classification task. And we provided interpretability by visualizing the decision-making process on different datasets.

In our proposed method, the use of a perfect binary tree may result in redundant nodes. Therefore, in future work, we can address this issue by applying pruning techniques to reduce redundancy. Pruning can simplify the model structure, improve computational efficiency, and enhance generalization capabilities. By pruning, we can also eliminate irrelevant features and nodes, making the decision-making process clearer and more interpretable.

\section*{Acknowledgements}
This work was supported by the National Natural Science Foundation of China under Grant 82072014, the Innovation Methods Work Special Project under Grant 2020IM020100, and the Natural Science Foundation of Shandong Province under Grant ZR2020QF112, ZR2020MF028. We would like to thank Eamonn Keogh and his team, Tony Bagnall and his team for the UEA/UCR time series classification repository.

\section*{Authors contributions}
Mingsen Du: Conceptualization, Methodology, Validation, Writing–original draft, Writing–review \& editing. Yanxuan Wei: Methodology, Validation, Writing–original draft, Writing–review \& editing. Yingxia Tang: Methodology, Validation, Writing–original draft, Writing–review \& editing. Xiangwei Zheng: Supervision, Project administration. Shoushui Wei: Methodology, Validation, Writing–original draft, Writing–review \& editing, Supervision, Project administration. Cun Ji: Methodology, Validation, Writing–original draft, Writing–review \& editing, Supervision, Project administration.

\section*{Data Availability} The datasets used or analyzed during the current study are available from the UEA archive: \url{http://timeseriesclassification.com}.

\bibliography{referencee}

\begin{thebibliography}{53}
\expandafter\ifx\csname natexlab\endcsname\relax\def\natexlab#1{#1}\fi
\providecommand{\url}[1]{\texttt{#1}}
\providecommand{\href}[2]{#2}
\providecommand{\path}[1]{#1}
\providecommand{\DOIprefix}{doi:}
\providecommand{\ArXivprefix}{arXiv:}
\providecommand{\URLprefix}{URL: }
\providecommand{\Pubmedprefix}{pmid:}
\providecommand{\doi}[1]{\href{http://dx.doi.org/#1}{\path{#1}}}
\providecommand{\Pubmed}[1]{\href{pmid:#1}{\path{#1}}}
\providecommand{\bibinfo}[2]{#2}
\ifx\xfnm\relax \def\xfnm[#1]{\unskip,\space#1}\fi
\bibitem[{Bahri et~al.(2022)Bahri, Li, Boubrahimi \&
  Hamdi}]{Bahri2022ShapeletbasedTA}
\bibinfo{author}{Bahri, O.~B.}, \bibinfo{author}{Li, P.},
  \bibinfo{author}{Boubrahimi, S.~F.}, \& \bibinfo{author}{Hamdi, S.~M.}
  (\bibinfo{year}{2022}).
\newblock \bibinfo{title}{Shapelet-based temporal association rule mining for
  multivariate time series classification}.
\newblock {\it \bibinfo{journal}{2022 IEEE International Conference on Big Data
  (Big Data)}\/},  (pp. \bibinfo{pages}{242--251}).
  \DOIprefix\doi{10.1109/BigData55660.2022.10020478}.
\bibitem[{Chambon et~al.(2017)Chambon, Galtier, Arnal, Wainrib \&
  Gramfort}]{Chambon2017ADL}
\bibinfo{author}{Chambon, S.}, \bibinfo{author}{Galtier, M.~N.},
  \bibinfo{author}{Arnal, P.~J.}, \bibinfo{author}{Wainrib, G.}, \&
  \bibinfo{author}{Gramfort, A.} (\bibinfo{year}{2017}).
\newblock \bibinfo{title}{A deep learning architecture for temporal sleep stage
  classification using multivariate and multimodal time series}.
\newblock {\it \bibinfo{journal}{IEEE Transactions on Neural Systems and
  Rehabilitation Engineering}\/},  {\it \bibinfo{volume}{26}\/},
  \bibinfo{pages}{758--769}. \DOIprefix\doi{10.1109/TNSRE.2018.2813138}.
\bibitem[{Chen et~al.(2018)Chen, Li, Barnett, Su \& Rudin}]{Chen2018ThisLL}
\bibinfo{author}{Chen, C.}, \bibinfo{author}{Li, O.}, \bibinfo{author}{Barnett,
  A.~J.}, \bibinfo{author}{Su, J.}, \& \bibinfo{author}{Rudin, C.}
  (\bibinfo{year}{2018}).
\newblock \bibinfo{title}{This looks like that: deep learning for interpretable
  image recognition}.
\newblock In {\it \bibinfo{booktitle}{Neural Information Processing
  Systems}\/}.
\bibitem[{Chen \& Wan(2023)}]{Chen2023LocalizedSS}
\bibinfo{author}{Chen, J.}, \& \bibinfo{author}{Wan, Y.}
  (\bibinfo{year}{2023}).
\newblock \bibinfo{title}{Localized shapelets selection for interpretable time
  series classification}.
\newblock {\it \bibinfo{journal}{Applied Intelligence}\/},  {\it
  \bibinfo{volume}{53}\/}, \bibinfo{pages}{17985--18001}.
  \DOIprefix\doi{10.1007/s10489-022-04422-2}.
\bibitem[{Chen \& Shi(2021)}]{CHEN2021126}
\bibinfo{author}{Chen, W.}, \& \bibinfo{author}{Shi, K.}
  (\bibinfo{year}{2021}).
\newblock \bibinfo{title}{Multi-scale attention convolutional neural network
  for time series classification}.
\newblock {\it \bibinfo{journal}{Neural Networks}\/},  {\it
  \bibinfo{volume}{136}\/}, \bibinfo{pages}{126--140}.
  \DOIprefix\doi{10.1016/j.neunet.2021.01.001}.
\bibitem[{Chen et~al.(2013)Chen, Hu, Keogh \& Batista}]{Chen2013DTWDTS}
\bibinfo{author}{Chen, Y.}, \bibinfo{author}{Hu, B.}, \bibinfo{author}{Keogh,
  E.~J.}, \& \bibinfo{author}{Batista, G. E. A. P.~A.} (\bibinfo{year}{2013}).
\newblock \bibinfo{title}{Dtw-d: time series semi-supervised learning from a
  single example}.
\newblock {\it \bibinfo{journal}{Proceedings of the 19th ACM SIGKDD
  international conference on Knowledge discovery and data mining}\/}, .
  \DOIprefix\doi{10.1145/2487575.2487633}.
\bibitem[{Cheng et~al.(2023)Cheng, Liu, Liu, Li, Luo \&
  Chen}]{Cheng2023FormerTimeHM}
\bibinfo{author}{Cheng, M.}, \bibinfo{author}{Liu, Q.}, \bibinfo{author}{Liu,
  Z.}, \bibinfo{author}{Li, Z.}, \bibinfo{author}{Luo, Y.}, \&
  \bibinfo{author}{Chen, E.} (\bibinfo{year}{2023}).
\newblock \bibinfo{title}{Formertime: Hierarchical multi-scale representations
  for multivariate time series classification}.
\newblock {\it \bibinfo{journal}{Proceedings of the ACM Web Conference
  2023}\/}, . \DOIprefix\doi{10.1145/3543507.3583205}.
\bibitem[{Delaney et~al.(2020)Delaney, Greene \&
  Keane}]{Delaney2020InstanceBasedCE}
\bibinfo{author}{Delaney, E.}, \bibinfo{author}{Greene, D.}, \&
  \bibinfo{author}{Keane, M.~T.} (\bibinfo{year}{2020}).
\newblock \bibinfo{title}{Instance-based counterfactual explanations for time
  series classification}.
\newblock {\it \bibinfo{journal}{ArXiv}\/},  {\it
  \bibinfo{volume}{abs/2009.13211}\/}.
  \DOIprefix\doi{10.1007/978-3-030-86957-1_3}.
\bibitem[{Dempster et~al.(2019)Dempster, Petitjean \&
  Webb}]{Dempster2019ROCKETEF}
\bibinfo{author}{Dempster, A.}, \bibinfo{author}{Petitjean, F.}, \&
  \bibinfo{author}{Webb, G.~I.} (\bibinfo{year}{2019}).
\newblock \bibinfo{title}{Rocket: exceptionally fast and accurate time series
  classification using random convolutional kernels}.
\newblock {\it \bibinfo{journal}{Data Mining and Knowledge Discovery}\/},  {\it
  \bibinfo{volume}{34}\/}, \bibinfo{pages}{1454 -- 1495}.
  \DOIprefix\doi{10.1007/s10618-020-00701-z}.
\bibitem[{Du et~al.(2023{\natexlab{a}})Du, Wei, Hu, Zheng \&
  Ji}]{Du2023AutomaticCI}
\bibinfo{author}{Du, M.}, \bibinfo{author}{Wei, Y.}, \bibinfo{author}{Hu, Y.},
  \bibinfo{author}{Zheng, X.}, \& \bibinfo{author}{Ji, C.}
  (\bibinfo{year}{2023}{\natexlab{a}}).
\newblock \bibinfo{title}{Automatic component identification based on time
  series classification for intelligent devices}.
\newblock {\it \bibinfo{journal}{2023 IEEE International Conference on Data
  Mining Workshops (ICDMW)}\/},  (pp. \bibinfo{pages}{540--549}).
  \DOIprefix\doi{10.1109/ICDMW60847.2023.00077}.
\bibitem[{Du et~al.(2024)Du, Wei, Hu, Zheng \& Ji}]{Du2024MultivariateTS}
\bibinfo{author}{Du, M.}, \bibinfo{author}{Wei, Y.}, \bibinfo{author}{Hu, Y.},
  \bibinfo{author}{Zheng, X.}, \& \bibinfo{author}{Ji, C.}
  (\bibinfo{year}{2024}).
\newblock \bibinfo{title}{Multivariate time series classification based on
  fusion features}.
\newblock {\it \bibinfo{journal}{Expert Systems with Applications}\/}, .
  \DOIprefix\doi{10.1016/j.eswa.2024.123452}.
\bibitem[{Du et~al.(2023{\natexlab{b}})Du, Wei, Zheng \&
  Ji}]{Du2023MultifeatureBN}
\bibinfo{author}{Du, M.}, \bibinfo{author}{Wei, Y.}, \bibinfo{author}{Zheng,
  X.}, \& \bibinfo{author}{Ji, C.} (\bibinfo{year}{2023}{\natexlab{b}}).
\newblock \bibinfo{title}{Multi-feature based network for multivariate time
  series classification}.
\newblock {\it \bibinfo{journal}{Inf. Sci.}\/},  {\it \bibinfo{volume}{639}\/},
  \bibinfo{pages}{119009}. \DOIprefix\doi{10.1016/j.ins.2023.119009}.
\bibitem[{Eldele et~al.(2021)Eldele, Ragab, Chen, Wu, Kwoh, Li \&
  Guan}]{Eldele2021TimeSeriesRL}
\bibinfo{author}{Eldele, E.}, \bibinfo{author}{Ragab, M.},
  \bibinfo{author}{Chen, Z.}, \bibinfo{author}{Wu, M.}, \bibinfo{author}{Kwoh,
  C.}, \bibinfo{author}{Li, X.}, \& \bibinfo{author}{Guan, C.}
  (\bibinfo{year}{2021}).
\newblock \bibinfo{title}{Time-series representation learning via temporal and
  contextual contrasting}.
\newblock In {\it \bibinfo{booktitle}{International Joint Conference on
  Artificial Intelligence}\/}.
\newblock \DOIprefix\doi{10.24963/ijcai.2021/324}.
\bibitem[{Fang et~al.(2018)Fang, Wang \& Wang}]{Fang2018EfficientLI}
\bibinfo{author}{Fang, Z.}, \bibinfo{author}{Wang, P.}, \&
  \bibinfo{author}{Wang, W.} (\bibinfo{year}{2018}).
\newblock \bibinfo{title}{Efficient learning interpretable shapelets for
  accurate time series classification}.
\newblock {\it \bibinfo{journal}{2018 IEEE 34th International Conference on
  Data Engineering (ICDE)}\/},  (pp. \bibinfo{pages}{497--508}).
  \DOIprefix\doi{10.1109/ICDE.2018.00052}.
\bibitem[{Franceschi et~al.(2019)Franceschi, Dieuleveut \&
  Jaggi}]{Franceschi2019UnsupervisedSR}
\bibinfo{author}{Franceschi, J.-Y.}, \bibinfo{author}{Dieuleveut, A.}, \&
  \bibinfo{author}{Jaggi, M.} (\bibinfo{year}{2019}).
\newblock \bibinfo{title}{Unsupervised scalable representation learning for
  multivariate time series}.
\newblock {\it \bibinfo{journal}{ArXiv}\/},  {\it
  \bibinfo{volume}{abs/1901.10738}\/}.
\bibitem[{Fu et~al.(2024)Fu, Li \& Shi}]{FU2024106234}
\bibinfo{author}{Fu, K.}, \bibinfo{author}{Li, H.}, \& \bibinfo{author}{Shi,
  X.} (\bibinfo{year}{2024}).
\newblock \bibinfo{title}{Ctf-former: A novel simplified multi-task learning
  strategy for simultaneous multivariate chaotic time series prediction}.
\newblock {\it \bibinfo{journal}{Neural Networks}\/},  {\it
  \bibinfo{volume}{174}\/}, \bibinfo{pages}{106234}.
  \DOIprefix\doi{10.1016/j.neunet.2024.106234}.
\bibitem[{Ghods \& Cook(2022)}]{Ghods2022PIPPI}
\bibinfo{author}{Ghods, A.}, \& \bibinfo{author}{Cook, D.~J.}
  (\bibinfo{year}{2022}).
\newblock \bibinfo{title}{Pip: Pictorial interpretable prototype learning for
  time series classification}.
\newblock {\it \bibinfo{journal}{IEEE Computational Intelligence Magazine}\/},
  {\it \bibinfo{volume}{17}\/}, \bibinfo{pages}{34--45}.
  \DOIprefix\doi{10.1109/MCI.2021.3129957}.
\bibitem[{H{\"o}llig et~al.(2022)H{\"o}llig, Kulbach \&
  Thoma}]{Hllig2022TSEvoEC}
\bibinfo{author}{H{\"o}llig, J.}, \bibinfo{author}{Kulbach, C.}, \&
  \bibinfo{author}{Thoma, S.} (\bibinfo{year}{2022}).
\newblock \bibinfo{title}{Tsevo: Evolutionary counterfactual explanations for
  time series classification}.
\newblock {\it \bibinfo{journal}{2022 21st IEEE International Conference on
  Machine Learning and Applications (ICMLA)}\/},  (pp.
  \bibinfo{pages}{29--36}). \DOIprefix\doi{10.1109/ICMLA55696.2022.00013}.
\bibitem[{Hou et~al.(2016)Hou, Kwok \& Zurada}]{Hou2016EfficientLO}
\bibinfo{author}{Hou, L.}, \bibinfo{author}{Kwok, J. T.-Y.}, \&
  \bibinfo{author}{Zurada, J.~M.} (\bibinfo{year}{2016}).
\newblock \bibinfo{title}{Efficient learning of timeseries shapelets}.
\newblock In {\it \bibinfo{booktitle}{AAAI Conference on Artificial
  Intelligence}\/}.
\newblock \DOIprefix\doi{10.1609/aaai.v30i1.10178}.
\bibitem[{Huang \& Deng(2023)}]{HUANG2023868}
\bibinfo{author}{Huang, F.}, \& \bibinfo{author}{Deng, Y.}
  (\bibinfo{year}{2023}).
\newblock \bibinfo{title}{Tcgan: Convolutional generative adversarial network
  for time series classification and clustering}.
\newblock {\it \bibinfo{journal}{Neural Networks}\/},  {\it
  \bibinfo{volume}{165}\/}, \bibinfo{pages}{868--883}.
  \DOIprefix\doi{10.1016/j.neunet.2023.06.033}.
\bibitem[{Ismail et~al.(2020)Ismail, Gunady, Bravo \&
  Feizi}]{Ismail2020BenchmarkingDL}
\bibinfo{author}{Ismail, A.~A.}, \bibinfo{author}{Gunady, M.~K.},
  \bibinfo{author}{Bravo, H.~C.}, \& \bibinfo{author}{Feizi, S.}
  (\bibinfo{year}{2020}).
\newblock \bibinfo{title}{Benchmarking deep learning interpretability in time
  series predictions}.
\newblock {\it \bibinfo{journal}{ArXiv}\/},  {\it
  \bibinfo{volume}{abs/2010.13924}\/}.
\bibitem[{Ji et~al.(2022{\natexlab{a}})Ji, Du, Hu, Liu, Pan \&
  Zheng}]{Ji2022TimeSC}
\bibinfo{author}{Ji, C.}, \bibinfo{author}{Du, M.}, \bibinfo{author}{Hu, Y.},
  \bibinfo{author}{Liu, S.}, \bibinfo{author}{Pan, L.}, \&
  \bibinfo{author}{Zheng, X.} (\bibinfo{year}{2022}{\natexlab{a}}).
\newblock \bibinfo{title}{Time series classification based on temporal
  features}.
\newblock {\it \bibinfo{journal}{Appl. Soft Comput.}\/},  {\it
  \bibinfo{volume}{128}\/}, \bibinfo{pages}{109494}.
  \DOIprefix\doi{10.1016/j.asoc.2022.109494}.
\bibitem[{Ji et~al.(2022{\natexlab{b}})Ji, Hu, Liu, Pan, Li \&
  Zheng}]{Ji2022FullyCN}
\bibinfo{author}{Ji, C.}, \bibinfo{author}{Hu, Y.}, \bibinfo{author}{Liu, S.},
  \bibinfo{author}{Pan, L.}, \bibinfo{author}{Li, B.}, \&
  \bibinfo{author}{Zheng, X.} (\bibinfo{year}{2022}{\natexlab{b}}).
\newblock \bibinfo{title}{Fully convolutional networks with shapelet features
  for time series classification}.
\newblock {\it \bibinfo{journal}{Inf. Sci.}\/},  {\it \bibinfo{volume}{612}\/},
  \bibinfo{pages}{835--847}. \DOIprefix\doi{10.1016/j.ins.2022.09.009}.
\bibitem[{Ji et~al.(2019)Ji, Zhao, Liu, Yang, Pan, Wu \& Meng}]{Ji2019AFS}
\bibinfo{author}{Ji, C.}, \bibinfo{author}{Zhao, C.}, \bibinfo{author}{Liu,
  S.}, \bibinfo{author}{Yang, C.}, \bibinfo{author}{Pan, L.},
  \bibinfo{author}{Wu, L.}, \& \bibinfo{author}{Meng, X.}
  (\bibinfo{year}{2019}).
\newblock \bibinfo{title}{A fast shapelet selection algorithm for time series
  classification}.
\newblock {\it \bibinfo{journal}{Comput. Networks}\/},  {\it
  \bibinfo{volume}{148}\/}, \bibinfo{pages}{231--240}.
  \DOIprefix\doi{10.1016/j.comnet.2018.11.031}.
\bibitem[{Karim et~al.(2018)Karim, Majumdar, Darabi \&
  Harford}]{Karim2018MultivariateLF}
\bibinfo{author}{Karim, F.}, \bibinfo{author}{Majumdar, S.},
  \bibinfo{author}{Darabi, H.}, \& \bibinfo{author}{Harford, S.}
  (\bibinfo{year}{2018}).
\newblock \bibinfo{title}{Multivariate lstm-fcns for time series
  classification}.
\newblock {\it \bibinfo{journal}{Neural networks : the official journal of the
  International Neural Network Society}\/},  {\it \bibinfo{volume}{116}\/},
  \bibinfo{pages}{237--245}. \DOIprefix\doi{10.1016/j.neunet.2019.04.014}.
\bibitem[{Kim et~al.(2022)Kim, Nam \& Ko}]{Kim2022ViTNeTIV}
\bibinfo{author}{Kim, S.}, \bibinfo{author}{Nam, J.~Y.}, \&
  \bibinfo{author}{Ko, B.} (\bibinfo{year}{2022}).
\newblock \bibinfo{title}{Vit-net: Interpretable vision transformers with
  neural tree decoder}.
\newblock In {\it \bibinfo{booktitle}{International Conference on Machine
  Learning}\/}.
\bibitem[{Lee et~al.(2023)Lee, Lindgren \& Papapetrou}]{Lee2023ZTimeEA}
\bibinfo{author}{Lee, Z.}, \bibinfo{author}{Lindgren, T.}, \&
  \bibinfo{author}{Papapetrou, P.} (\bibinfo{year}{2023}).
\newblock \bibinfo{title}{Z-time: efficient and effective interpretable
  multivariate time series classification}.
\newblock {\it \bibinfo{journal}{Data Mining and Knowledge Discovery}\/},  (pp.
  \bibinfo{pages}{1--31}). \DOIprefix\doi{10.1007/s10618-023-00969-x}.
\bibitem[{Li et~al.(2021)Li, Choi, Xu, Bhowmick, Chun \&
  Wong}]{Li2021ShapeNetAS}
\bibinfo{author}{Li, G.}, \bibinfo{author}{Choi, B.}, \bibinfo{author}{Xu, J.},
  \bibinfo{author}{Bhowmick, S.~S.}, \bibinfo{author}{Chun, K.-P.}, \&
  \bibinfo{author}{Wong, G.~L.} (\bibinfo{year}{2021}).
\newblock \bibinfo{title}{Shapenet: A shapelet-neural network approach for
  multivariate time series classification}.
\newblock In {\it \bibinfo{booktitle}{AAAI Conference on Artificial
  Intelligence}\/}.
\newblock \DOIprefix\doi{10.1609/aaai.v35i9.17018}.
\bibitem[{Li et~al.(2023)Li, Zhang, Zhu, Liu, Han, Zhang \&
  Wei}]{li2023diagnosis}
\bibinfo{author}{Li, Y.}, \bibinfo{author}{Zhang, L.}, \bibinfo{author}{Zhu,
  L.}, \bibinfo{author}{Liu, L.}, \bibinfo{author}{Han, B.},
  \bibinfo{author}{Zhang, Y.}, \& \bibinfo{author}{Wei, S.}
  (\bibinfo{year}{2023}).
\newblock \bibinfo{title}{Diagnosis of atrial fibrillation using
  self-complementary attentional convolutional neural network}.
\newblock {\it \bibinfo{journal}{Computer Methods and Programs in
  Biomedicine}\/},  {\it \bibinfo{volume}{238}\/}, \bibinfo{pages}{107565}.
\bibitem[{Li et~al.(2015)Li, Jin \& Zhao}]{Li2015DrunkDD}
\bibinfo{author}{Li, Z.}, \bibinfo{author}{Jin, X.}, \& \bibinfo{author}{Zhao,
  X.} (\bibinfo{year}{2015}).
\newblock \bibinfo{title}{Drunk driving detection based on classification of
  multivariate time series.}
\newblock {\it \bibinfo{journal}{Journal of safety research}\/},  {\it
  \bibinfo{volume}{54}\/}, \bibinfo{pages}{61--4}.
  \DOIprefix\doi{10.1016/j.jsr.2015.06.007}.
\bibitem[{Liu et~al.(2021)Liu, Lin, Cao, Hu, Wei, Zhang, Lin \&
  Guo}]{Liu2021SwinTH}
\bibinfo{author}{Liu, Z.}, \bibinfo{author}{Lin, Y.}, \bibinfo{author}{Cao,
  Y.}, \bibinfo{author}{Hu, H.}, \bibinfo{author}{Wei, Y.},
  \bibinfo{author}{Zhang, Z.}, \bibinfo{author}{Lin, S.}, \&
  \bibinfo{author}{Guo, B.} (\bibinfo{year}{2021}).
\newblock \bibinfo{title}{Swin transformer: Hierarchical vision transformer
  using shifted windows}.
\newblock {\it \bibinfo{journal}{2021 IEEE/CVF International Conference on
  Computer Vision (ICCV)}\/},  (pp. \bibinfo{pages}{9992--10002}).
  \DOIprefix\doi{10.1109/ICCV48922.2021.00986}.
\bibitem[{Ma et~al.(2019)Ma, Li, Zhang, Gao \& Lu}]{Ma2019AttnSenseMA}
\bibinfo{author}{Ma, H.}, \bibinfo{author}{Li, W.}, \bibinfo{author}{Zhang,
  X.}, \bibinfo{author}{Gao, S.}, \& \bibinfo{author}{Lu, S.}
  (\bibinfo{year}{2019}).
\newblock \bibinfo{title}{Attnsense: Multi-level attention mechanism for
  multimodal human activity recognition}.
\newblock In {\it \bibinfo{booktitle}{International Joint Conference on
  Artificial Intelligence}\/}.
\newblock \DOIprefix\doi{10.24963/ijcai.2019/431}.
\bibitem[{Ma et~al.(2020)Ma, Zhuang, Li, Huang \&
  Cottrell}]{Ma2020AdversarialDS}
\bibinfo{author}{Ma, Q.}, \bibinfo{author}{Zhuang, W.}, \bibinfo{author}{Li,
  S.}, \bibinfo{author}{Huang, D.}, \& \bibinfo{author}{Cottrell, G.}
  (\bibinfo{year}{2020}).
\newblock \bibinfo{title}{Adversarial dynamic shapelet networks}.
\newblock In {\it \bibinfo{booktitle}{AAAI Conference on Artificial
  Intelligence}\/}.
\newblock \DOIprefix\doi{10.1609/AAAI.V34I04.5948}.
\bibitem[{Manzella et~al.(2021)Manzella, Pagliarini, Sciavicco \&
  Stan}]{Manzella2021IntervalTR}
\bibinfo{author}{Manzella, F.}, \bibinfo{author}{Pagliarini, G.},
  \bibinfo{author}{Sciavicco, G.}, \& \bibinfo{author}{Stan, E.~I.}
  (\bibinfo{year}{2021}).
\newblock \bibinfo{title}{Interval temporal random forests with an application
  to covid-19 diagnosis}.
\newblock In {\it \bibinfo{booktitle}{Time}\/}.
\newblock \DOIprefix\doi{10.4230/LIPIcs.TIME.2021.7}.
\bibitem[{Middlehurst et~al.(2019)Middlehurst, Vickers \&
  Bagnall}]{Middlehurst2019ScalableDC}
\bibinfo{author}{Middlehurst, M.}, \bibinfo{author}{Vickers, W.~D.}, \&
  \bibinfo{author}{Bagnall, A.} (\bibinfo{year}{2019}).
\newblock \bibinfo{title}{Scalable dictionary classifiers for time series
  classification}.
\newblock {\it \bibinfo{journal}{ArXiv}\/},  {\it
  \bibinfo{volume}{abs/1907.11815}\/}.
  \DOIprefix\doi{10.1007/978-3-030-33607-3_2}.
\bibitem[{Nguyen et~al.(2019)Nguyen, Gsponer, Ilie, O'Reilly \&
  Ifrim}]{Nguyen2019InterpretableTS}
\bibinfo{author}{Nguyen, T.~L.}, \bibinfo{author}{Gsponer, S.},
  \bibinfo{author}{Ilie, I.}, \bibinfo{author}{O'Reilly, M.}, \&
  \bibinfo{author}{Ifrim, G.} (\bibinfo{year}{2019}).
\newblock \bibinfo{title}{Interpretable time series classification using linear
  models and multi-resolution multi-domain symbolic representations}.
\newblock {\it \bibinfo{journal}{Data Mining and Knowledge Discovery}\/},  {\it
  \bibinfo{volume}{33}\/}, \bibinfo{pages}{1183 -- 1222}.
  \DOIprefix\doi{10.1007/s10618-019-00633-3}.
\bibitem[{Pagliarini et~al.(2022)Pagliarini, Scaboro, Serra, Sciavicco \&
  Stan}]{Pagliarini2022NeuralSymbolicTD}
\bibinfo{author}{Pagliarini, G.}, \bibinfo{author}{Scaboro, S.},
  \bibinfo{author}{Serra, G.}, \bibinfo{author}{Sciavicco, G.}, \&
  \bibinfo{author}{Stan, E.~I.} (\bibinfo{year}{2022}).
\newblock \bibinfo{title}{Neural-symbolic temporal decision trees for
  multivariate time series classification}.
\newblock In {\it \bibinfo{booktitle}{Time}\/}.
\newblock \DOIprefix\doi{10.4230/LIPIcs.TIME.2022.13}.
\bibitem[{Rojat et~al.(2021)Rojat, Puget, Filliat, Ser, Gelin \&
  D'iaz-Rodr'iguez}]{Rojat2021ExplainableAI}
\bibinfo{author}{Rojat, T.}, \bibinfo{author}{Puget, R.},
  \bibinfo{author}{Filliat, D.}, \bibinfo{author}{Ser, J.~D.},
  \bibinfo{author}{Gelin, R.}, \& \bibinfo{author}{D'iaz-Rodr'iguez, N.}
  (\bibinfo{year}{2021}).
\newblock \bibinfo{title}{Explainable artificial intelligence (xai) on
  timeseries data: A survey}.
\newblock {\it \bibinfo{journal}{ArXiv}\/},  {\it
  \bibinfo{volume}{abs/2104.00950}\/}.
\bibitem[{Ruiz et~al.(2020)Ruiz, Flynn, Large, Middlehurst \&
  Bagnall}]{Ruiz2020TheGM}
\bibinfo{author}{Ruiz, A.~P.}, \bibinfo{author}{Flynn, M.},
  \bibinfo{author}{Large, J.}, \bibinfo{author}{Middlehurst, M.}, \&
  \bibinfo{author}{Bagnall, A.} (\bibinfo{year}{2020}).
\newblock \bibinfo{title}{The great multivariate time series classification
  bake off: a review and experimental evaluation of recent algorithmic
  advances}.
\newblock {\it \bibinfo{journal}{Data Mining and Knowledge Discovery}\/},  {\it
  \bibinfo{volume}{35}\/}, \bibinfo{pages}{401 -- 449}.
  \DOIprefix\doi{10.1007/s10618-020-00727-3}.
\bibitem[{Tang et~al.(2019)Tang, Liu \& Long}]{Tang2019FewshotTC}
\bibinfo{author}{Tang, W.}, \bibinfo{author}{Liu, L.}, \&
  \bibinfo{author}{Long, G.} (\bibinfo{year}{2019}).
\newblock \bibinfo{title}{Few-shot time-series classiﬁcation with dual
  interpretability}.
\newblock \URLprefix \url{https://api.semanticscholar.org/CorpusID:211167393}.
\bibitem[{Tao et~al.(2022)Tao, Yang \& Jing}]{Tao2022ProfilematrixbasedSD}
\bibinfo{author}{Tao, Q.}, \bibinfo{author}{Yang, J.}, \&
  \bibinfo{author}{Jing, S.} (\bibinfo{year}{2022}).
\newblock \bibinfo{title}{Profile-matrix-based shapelet discovery for time
  series binary classification}.
\newblock {\it \bibinfo{journal}{2022 18th International Conference on
  Computational Intelligence and Security (CIS)}\/},  (pp.
  \bibinfo{pages}{297--301}). \DOIprefix\doi{10.1109/CIS58238.2022.00069}.
\bibitem[{Theissler et~al.(2022)Theissler, Spinnato, Schlegel \&
  Guidotti}]{Theissler2022ExplainableAF}
\bibinfo{author}{Theissler, A.}, \bibinfo{author}{Spinnato, F.},
  \bibinfo{author}{Schlegel, U.}, \& \bibinfo{author}{Guidotti, R.}
  (\bibinfo{year}{2022}).
\newblock \bibinfo{title}{Explainable ai for time series classification: A
  review, taxonomy and research directions}.
\newblock {\it \bibinfo{journal}{IEEE Access}\/},  {\it
  \bibinfo{volume}{10}\/}, \bibinfo{pages}{100700--100724}.
  \DOIprefix\doi{10.1109/ACCESS.2022.3207765}.
\bibitem[{Tonekaboni et~al.(2021)Tonekaboni, Eytan \&
  Goldenberg}]{Tonekaboni2021UnsupervisedRL}
\bibinfo{author}{Tonekaboni, S.}, \bibinfo{author}{Eytan, D.}, \&
  \bibinfo{author}{Goldenberg, A.} (\bibinfo{year}{2021}).
\newblock \bibinfo{title}{Unsupervised representation learning for time series
  with temporal neighborhood coding}.
\newblock {\it \bibinfo{journal}{ArXiv}\/},  {\it
  \bibinfo{volume}{abs/2106.00750}\/}.
\bibitem[{Turb'e et~al.(2022)Turb'e, Bjelogrlic, Lovis \&
  Mengaldo}]{Turbe2022InterpretTimeAN}
\bibinfo{author}{Turb'e, H.}, \bibinfo{author}{Bjelogrlic, M.},
  \bibinfo{author}{Lovis, C.}, \& \bibinfo{author}{Mengaldo, G.}
  (\bibinfo{year}{2022}).
\newblock \bibinfo{title}{Interprettime: a new approach for the systematic
  evaluation of neural-network interpretability in time series classification}.
\newblock {\it \bibinfo{journal}{ArXiv}\/},  {\it
  \bibinfo{volume}{abs/2202.05656}\/}.
\bibitem[{Wan et~al.(2024)Wan, Xia, Yin, Pan, Hu \& Yi}]{WAN2024106196}
\bibinfo{author}{Wan, J.}, \bibinfo{author}{Xia, N.}, \bibinfo{author}{Yin,
  Y.}, \bibinfo{author}{Pan, X.}, \bibinfo{author}{Hu, J.}, \&
  \bibinfo{author}{Yi, J.} (\bibinfo{year}{2024}).
\newblock \bibinfo{title}{Tcdformer: A transformer framework for non-stationary
  time series forecasting based on trend and change-point detection}.
\newblock {\it \bibinfo{journal}{Neural Networks}\/},  {\it
  \bibinfo{volume}{173}\/}, \bibinfo{pages}{106196}.
  \DOIprefix\doi{10.1016/j.neunet.2024.106196}.
\bibitem[{Wan et~al.(2023)Wan, Cen, Chen, Xie \& Gui}]{Wan2023MemorySL}
\bibinfo{author}{Wan, X.}, \bibinfo{author}{Cen, L.}, \bibinfo{author}{Chen,
  X.}, \bibinfo{author}{Xie, Y.}, \& \bibinfo{author}{Gui, W.}
  (\bibinfo{year}{2023}).
\newblock \bibinfo{title}{Memory shapelet learning for early classification of
  streaming time series.}
\newblock {\it \bibinfo{journal}{IEEE transactions on cybernetics}\/},  {\it
  \bibinfo{volume}{PP}\/}. \DOIprefix\doi{10.1109/TCYB.2023.3337550}.
\bibitem[{Wang et~al.(2016)Wang, Yan \& Oates}]{Wang2016TimeSC}
\bibinfo{author}{Wang, Z.}, \bibinfo{author}{Yan, W.}, \&
  \bibinfo{author}{Oates, T.} (\bibinfo{year}{2016}).
\newblock \bibinfo{title}{Time series classification from scratch with deep
  neural networks: A strong baseline}.
\newblock {\it \bibinfo{journal}{2017 International Joint Conference on Neural
  Networks (IJCNN)}\/},  (pp. \bibinfo{pages}{1578--1585}).
  \DOIprefix\doi{10.1109/IJCNN.2017.7966039}.
\bibitem[{Yang et~al.(2023)Yang, Wang, Yao, Long \& Xu}]{Yang2023DyformerAD}
\bibinfo{author}{Yang, C.}, \bibinfo{author}{Wang, X.}, \bibinfo{author}{Yao,
  L.}, \bibinfo{author}{Long, G.}, \& \bibinfo{author}{Xu, G.}
  (\bibinfo{year}{2023}).
\newblock \bibinfo{title}{Dyformer: A dynamic transformer-based architecture
  for multivariate time series classification}.
\newblock {\it \bibinfo{journal}{Inf. Sci.}\/},  {\it \bibinfo{volume}{656}\/},
  \bibinfo{pages}{119881}. \DOIprefix\doi{10.2139/ssrn.4362666}.
\bibitem[{Younis et~al.(2023)Younis, Ahmadi, Hakmeh \&
  Fisichella}]{Younis2023FLAMES2GraphAI}
\bibinfo{author}{Younis, R.}, \bibinfo{author}{Ahmadi, Z.},
  \bibinfo{author}{Hakmeh, A.}, \& \bibinfo{author}{Fisichella, M.}
  (\bibinfo{year}{2023}).
\newblock \bibinfo{title}{Flames2graph: An interpretable federated multivariate
  time series classification framework}.
\newblock {\it \bibinfo{journal}{Proceedings of the 29th ACM SIGKDD Conference
  on Knowledge Discovery and Data Mining}\/}, .
  \DOIprefix\doi{10.1145/3580305.3599354}.
\bibitem[{Yu et~al.(2022)Yu, Zeng, Xue \& Ma}]{Yu2022LSTMBasedID}
\bibinfo{author}{Yu, Y.}, \bibinfo{author}{Zeng, X.}, \bibinfo{author}{Xue,
  X.}, \& \bibinfo{author}{Ma, J.} (\bibinfo{year}{2022}).
\newblock \bibinfo{title}{Lstm-based intrusion detection system for vanets: A
  time series classification approach to false message detection}.
\newblock {\it \bibinfo{journal}{IEEE Transactions on Intelligent
  Transportation Systems}\/},  {\it \bibinfo{volume}{23}\/},
  \bibinfo{pages}{23906--23918}. \DOIprefix\doi{10.1109/TITS.2022.3190432}.
\bibitem[{Zerveas et~al.(2020)Zerveas, Jayaraman, Patel, Bhamidipaty \&
  Eickhoff}]{Zerveas2020ATF}
\bibinfo{author}{Zerveas, G.}, \bibinfo{author}{Jayaraman, S.},
  \bibinfo{author}{Patel, D.}, \bibinfo{author}{Bhamidipaty, A.}, \&
  \bibinfo{author}{Eickhoff, C.} (\bibinfo{year}{2020}).
\newblock \bibinfo{title}{A transformer-based framework for multivariate time
  series representation learning}.
\newblock {\it \bibinfo{journal}{Proceedings of the 27th ACM SIGKDD Conference
  on Knowledge Discovery \& Data Mining}\/}, .
  \DOIprefix\doi{10.1145/3447548.3467401}.
\bibitem[{Zhu \& Hill(2021)}]{Zhu2021NetworkedTS}
\bibinfo{author}{Zhu, L.}, \& \bibinfo{author}{Hill, D.}
  (\bibinfo{year}{2021}).
\newblock \bibinfo{title}{Networked time series shapelet learning for power
  system transient stability assessment}.
\newblock {\it \bibinfo{journal}{IEEE Transactions on Power Systems}\/},  {\it
  \bibinfo{volume}{37}\/}, \bibinfo{pages}{416--428}.
  \DOIprefix\doi{10.1109/tpwrs.2021.3093423}.
\bibitem[{Zuo et~al.(2023)Zuo, Li, Choi, Bhowmick, yin Mah \&
  Wong}]{Zuo2023SVPTAS}
\bibinfo{author}{Zuo, R.}, \bibinfo{author}{Li, G.}, \bibinfo{author}{Choi,
  B.}, \bibinfo{author}{Bhowmick, S.~S.}, \bibinfo{author}{yin Mah, D.~N.}, \&
  \bibinfo{author}{Wong, G.~L.} (\bibinfo{year}{2023}).
\newblock \bibinfo{title}{Svp-t: A shape-level variable-position transformer
  for multivariate time series classification}.
\newblock In {\it \bibinfo{booktitle}{AAAI Conference on Artificial
  Intelligence}\/}.
\newblock \DOIprefix\doi{10.1609/aaai.v37i9.26359}.

\end{thebibliography}
\end{document}